\documentclass{ieeeaccess}
\usepackage{cite}
\usepackage{amsmath,amssymb,amsfonts}
\usepackage{algorithmic}
\usepackage{graphicx}
\usepackage{textcomp}

\usepackage{booktabs}
\usepackage{multirow}
\usepackage{hhline}
\usepackage{tabularray}
\usepackage{caption}
\usepackage{subcaption} 
\usepackage{hyperref}
\usepackage{balance}
\usepackage{xcolor}
\usepackage{bbding}
\newcommand{\etal}{\textit{et al.}}
\def\BibTeX{{\rm B\kern-.05em{\sc i\kern-.025em b}\kern-.08em
    T\kern-.1667em\lower.7ex\hbox{E}\kern-.125emX}}
\begin{document}
\history{Date of publication xxxx 00, 0000, date of current version xxxx 00, 0000.}
\doi{10.1109/ACCESS.2025.3528633}

\title{IncSAR: A Dual Fusion Incremental Learning Framework for SAR Target Recognition}
\author{\uppercase{George Karantaidis},
\uppercase{Athanasios Pantsios, \uppercase{Ioannis Kompatsiaris} \IEEEmembership{Senior Member, IEEE}, and Symeon Papadopoulos}}
\address{Centre for Research \& Technology Hellas, 570 01 Thessaloniki, Greece}

\tfootnote{This project has received funding from the FaRADAI project (ref. 101103386) funded by the European Commission under the European Defence Fund. Views and opinions expressed are however those of the author(s) only and do not necessarily reflect those of the European Union or the European Commission. Neither the European Union nor the granting authority can be held responsible for them.}

\markboth
{Author \headeretal: Preparation of Papers for IEEE TRANSACTIONS and JOURNALS}
{Author \headeretal: Preparation of Papers for IEEE TRANSACTIONS and JOURNALS}

\corresp{Corresponding author: George Karantaidis (e-mail: karantai@iti.gr).}

\begin{abstract}
Deep learning techniques have achieved significant success in Synthetic Aperture Radar (SAR) target recognition using predefined datasets in static scenarios. However, real-world applications demand that models incrementally learn new information without forgetting previously acquired knowledge. The challenge of catastrophic forgetting, where models lose past knowledge when adapting to new tasks, remains a critical issue. In this paper, we introduce IncSAR, an incremental learning framework designed to tackle catastrophic forgetting in SAR target recognition. IncSAR combines the power of a Vision Transformer (ViT) and a custom-designed Convolutional Neural Network (CNN) in a dual-branch architecture, integrated via a late-fusion strategy. Additionally, we explore the use of TinyViT to reduce computational complexity and propose an attention mechanism to dynamically enhance feature representation. To mitigate the speckle noise inherent in SAR images, we employ a denoising module based on a neural network approximation of Robust Principal Component Analysis (RPCA), leveraging a simple  neural network for efficient noise reduction in SAR imagery.  Moreover, a random projection layer improves the linear separability of features, and a variant of Linear Discriminant Analysis (LDA) decorrelates extracted class prototypes for better generalization. Extensive experiments on the MSTAR, SAR-AIRcraft-1.0, and OpenSARShip benchmark datasets demonstrate that IncSAR significantly outperforms state-of-the-art approaches, achieving a 99.63\% average accuracy and a 0.33\% performance drop, representing an 89\% improvement in retention compared to existing techniques. The source code is available at \url{https://github.com/geokarant/IncSAR}.
\end{abstract}


\titlepgskip=-15pt

\maketitle

\section{Introduction}
\label{sec:introduction}
\PARstart{S}{ynthetic} aperture radar (SAR) is an active remote sensing technology that obtains high-resolution images with minimal dependence on light, weather, and other environmental conditions. SAR automatic target recognition (SAR-ATR) through deep learning finds applications in a wide range of fields, such as target acquisition, disaster management, and maritime vigilance \cite{passah2022sar}. The interpretation of SAR images is considered to be a challenging task due to the presence of speckle noise. In contrast to optical images, SAR images tend to exhibit smaller inter-class and larger intra-class distances, rendering their classification particularly challenging \cite{li2023comprehensive}.

In practical settings, applications often deal with streaming data with incoming new classes that cannot be stored and recalled due to bounded storage or privacy issues. 
An additional challenge present in practical scenarios concerns data distribution shifts over time. Class incremental learning (CIL) aims to build models that continually adapt to new sets of classes while performing well among all seen classes. Catastrophic forgetting \cite{mccloskey1989catastrophic}, a crucial issue present in incremental learning, refers to the phenomenon where a model's performance on previously learned tasks deteriorates as it acquires new knowledge. A relevant challenge in CIL involves the stability-plasticity trade-off \cite{wu2021striking}, which refers to the balance between a model's ability to preserve old knowledge and its ability to adapt to new classes. Despite recent advancements in CIL methods, their performance remains significantly lower compared to conventional machine learning scenarios, especially in the face of an increasing number of incremental tasks.

One of the most popular CIL techniques includes regularization-based methods \cite{rebuffi2017icarl, hou2019learning}, which use regularization terms and typically involve storing a frozen copy of the old model, imposing constraints on important weights, or implementing knowledge distillation. Another category comprises parameter-isolation methods \cite{buzzega2020dark}, which modify or add network parameters or sub-modules according to task-specific requirements in order to adapt the network architecture during training to new tasks. Replay-based methods \cite{wang2024comprehensive} store or generate samples or representations from previous data to mitigate catastrophic forgetting. Exemplar-based methods \cite{wang2022foster}, a subset of replay methods, specifically require a rehearsal buffer to store a fixed number of samples from previous classes. In contrast, class-prototype based methods are exemplar-free methods that utilize a network for feature extraction and memorize a set of representative prototypes for each class, which are employed for classification purposes \cite{mcdonnell2024ranpac}. Recently, pre-trained models (PTMs), such as Vision Transformers (ViT) \cite{dosovitskiy2020image}, have demonstrated remarkable progress in generating strong representations, rendering them a good choice for CIL scenarios \cite{zhou2024continual}. The generalization capability of PTMs can be combined with parameter-efficient fine-tuning (PEFT) techniques to tackle the different distribution in downstream tasks \cite{burghouts2024synthesizing}.

Current incremental learning techniques for SAR-ATR deal mainly with specific challenges like catastrophic forgetting, speckle noise reduction, and high computational requirements, but struggle with issues that come up in real-world cases. Speckle noise, which deteriorates image quality and interferes with feature extraction, is a SAR-specific challenge that is often ignored by techniques that attempt to mitigate catastrophic forgetting. Additionally, rich feature extraction from SAR data is essential for precise classification, and current approaches are often limited in the level of feature information they extract (i.e., focus on either global or fine-grained). These limitations demonstrate the need for a comprehensive incremental learning framework for SAR-ATR images that is effective in a wide range of changing real-world circumstances.

To this end, we propose a class-prototype based incremental learning framework, termed IncSAR, for SAR target recognition.  IncSAR integrates a dual-branch architecture, combining a pre-trained ViT and a custom  Convolutional Neural Network (CNN) to ensure robust feature extraction. It incorporates robust PCA-based denoising to mitigate speckle noise, random projection layers to enhance feature separability, and a class-prototype learning approach that avoids rehearsal buffer reliance. Extensive experimental results have shown that IncSAR achieves state-of-the-art results, outperforming competitive approaches. These results demonstrate its ability to address challenges effectively, while deriving noteworthy performance in real-world incremental learning scenarios.

Specifically, we argue that PTMs can be successfully used in CIL for SAR target recognition reducing time requirements, enabling generalization to new tasks and cross-domain adaptation. IncSAR utilizes a pre-trained ViT and a custom-designed CNN, called SAR-CNN, as strong feature extractors, combining them in a late-fusion strategy to take advantage of their complementary strengths. A scale and shift method is employed for the PEFT of the PTM to mitigate the distribution mismatch problem in the downstream dataset. A CNN-based Robust Principal Component Analysis \cite{rpca, han2021efficient, feng2013online} is employed for noise despeckling prior the CNN feature extraction; a bilinear neural network is employed to derive a low-rank and a sparse component of the input SAR images. The extracted features are randomly projected in a higher-dimensional space to enhance the linear separability, and then they are utilized to extract the class prototypes. A linear discriminant analysis (LDA) \cite{panos2023first} approach is used for the decorrelation of prototypes, which are used for classification. Moreover, an attention mechanism is introduced within the IncSAR framework for feature fusion, resulting in improved SAR target classification.  Our main contributions are summarized as follows:
\begin{itemize}
    \item We propose the IncSAR framework, introducing a late-fusion strategy that combines a pre-trained ViT and a custom-designed CNN as network backbones. Both the  ViT-B/16 and a smaller variant, TinyViT \cite{wu2022tinyvit}, are employed in separate experiments to explore the trade-offs between model complexity and performance. The ViT models are fine-tuned using a scale-and-shift  method for PEFT.
    \item IncSAR adopts an exemplar-free prototype learning approach, eliminating the need for a rehearsal buffer. A variant of LDA is used to decorrelate the extracted class prototypes, improving the framework’s discriminative ability.
    \item An attention module, built on the 4-layer ViT-Ti architecture \cite{touvron2021training}, is incorporated into the IncSAR framework to enhance feature extraction, focusing on relevant patterns and improving SAR target classification.
    \item A bilinear network approximation of Robust PCA is employed for noise despeckling in SAR imagery further enhancing the classification accuracy of IncSAR.
    \item Extensive experiments on the MSTAR dataset demonstrate notable gains over state-of-the-art  approaches, achieving accuracies of 99.63\% and performance dropping rate improvement of 89\% compared to state-of-the-art. Additional experiments on the SAR-AIRcraft-1.0 dataset demonstrate the model's effectiveness in handling complex real-world scenarios. Moreover, IncSAR's generalization ability is evaluated using the OpenSARShip dataset, and ablation studies further attest to its robustness and efficiency.   
\end{itemize}

\begin{table*}[htb]
    \caption{Summary of SAR-ATR incremental methods. The table lists various incremental learning methods along with the datasets employed in each study. The datasets include SAR-specific imagery datasets such as MSTAR \cite{ross1998standard}, OpenSARShip \cite{huang2017opensarship}, SAMPLE \cite{lewis2019sar}, and SAR-Aircraft-1.0 \cite{zhirui2023sar}, as well as CIFAR-100 \cite{krizhevsky2009learning}, which, unlike the others, does not consist of SAR images but is used for benchmarking in computer vision tasks.}
    \label{tab:summary_sar}
    \centering
    \resizebox{2\columnwidth}{!}{%
        \begin{tabular}{l cccccc}
        \hline
            \multirow{2}{*}{Methods} & \multirow{2}{*}{Regularization}  & Replay/ & Parameter& Feature& \multirow{2}{*}{Dataset} & \multirow{2}{*}{Year}\\
        &&Exemplars&Isolation& Extractor&&\\
        \hline
        MEDIL  \cite{huang2023incremental} &\Checkmark   &\Checkmark && -&  MSTAR, OpenSARShip  & 2023  \\
        CBesIL \cite{dang2020class}  &  &\Checkmark && -&   MSTAR& 2020\\
        Zhou \etal \cite{zhou2022sar} & \Checkmark &\Checkmark&\Checkmark&ResNet-18 \cite{he2016deep} & MSTAR & 2022\\
        DCBES \cite{li2023density} &   &\Checkmark &  & CNN \cite{liu2019spotlight} &  MSTAR& 2023\\
        HPecIL \cite{tang2022incremental} & \Checkmark & \Checkmark & \Checkmark & ResNet-18 &  MSTAR&2022 \\                     
        Hu \etal \cite{hu2022incremental} & \Checkmark & \Checkmark& & Alexnet \cite{krizhevsky2012imagenet}& MSTAR & 2022\\
        ICAC  \cite{li2022incremental} & \Checkmark& \Checkmark &&CNN &  MSTAR, OpenSARShip & 2022\\
        MLAKDN \cite{yu2023multilevel} & \Checkmark & \Checkmark &&ResNet-18 &  MSTAR, SAMPLE& 2023\\
        DERDN \cite{ren2024dynamic} & \Checkmark & \Checkmark && ODConv \cite{li2022omni} &MSTAR, SAMPLE &2024 \\
        SSF-IL \cite{gao2024sar} &  & \Checkmark & & ResNet-18& MSTAR & 2024\\
        Pan \etal \cite{pan2023class} & \Checkmark & \Checkmark & \Checkmark & ViT\cite{dosovitskiy2020image}& MSTAR, CIFAR100 & 2023 \\
        CIL-MMI \cite{li2024sar} &\Checkmark & &&ResNet-18 &   MSTAR &2024\\
        IncSAR &\Checkmark & && ViT, SAR-CNN &   MSTAR, OpenSARShip, SAR-AIRcraft-1.0 &2024\\
        \hline
        \end{tabular}%
    } 
\end{table*}

\section{Related work}
\label{sec: related work}

\textbf{Class incremental learning}: CIL methods can be broadly divided into \cite{qiang2024fett}: regularization-based methods (iCaRL \cite{rebuffi2017icarl}, LUCIR \cite{hou2019learning}, Foster \cite{wang2022foster}), parameter-isolation based methods (DER \cite{buzzega2020dark}), replay-based methods (Fetril \cite{petit2023fetril}), and pre-trained methods \cite{zhou2024continual}. Recent studies focus heavily on pre-trained methods benefiting from the powerful feature extraction capabilities of PTMs, and mainly include prompt-based methods, class-prototype based methods, and model-mixture based methods. Prompt-based methods insert a small number of learnable parameters rather than fully fine-tuning the PTM's weights (L2P \cite{wang2022learning}, Coda-prompt \cite{smith2023coda}). Class-prototype based methods extract representative prototypes for each class and utilize them for classification (Adam \cite{zhou2023revisiting}, RanPAC \cite{mcdonnell2024ranpac}, SLCA \cite{zhang2023slca}). These methods can employ a frozen PTM or be combined with PEFT techniques, and they can also utilize prototype decorrelation techniques. The main idea of model-mixture based methods involves ensembling or merging various fine-tuned PTMs into a single model that integrates the representational capabilities of multiple models (PROOF \cite{zhou2023learning}, SEED \cite{rypesc2024divide}, CoFiMA \cite{marouf2023weighted}). 
These methods are highly complementary and can combine different approaches, depending on the priorities of the learning scenario.

\textbf{Class incremental learning for SAR-ATR}: Most existing methods for CIL in SAR-ATR are exemplar-based and rely on a bounded subset of past training data. A weight correction method, named MEDIL, was proposed in \cite{huang2023incremental} that utilizes a hybrid loss function to strike an optimal plasticity-stability trade-off. The CBesIL approach \cite{dang2020class} introduced a class-boundary selection method using local geometry and statistics, along with a resampling method for data distribution reconstruction. A major issue with replay methods concerns the imbalance between old and new classes due to the limited amount of old class data stored in the rehearsal buffer. Zhou \etal \cite{zhou2022sar} proposed a bias-correction layer to tackle the class imbalance problem. The process of selecting exemplars is critical in data replay methods. DCBES \cite{li2023density} utilized a greedy algorithm to select representative exemplar samples based on their density in the feature space. Tang \etal \cite{tang2022incremental} proposed a method named HPecIL, that combines replay and weight regularization techniques. HPecIL preserved multiple optimal models from old data, employing a pruning initialization method to remove low-impact nodes of the neural network, and using class-balanced training batches to address the distribution shift in the incremental tasks.  Hu \etal \cite{hu2022incremental} proposed the addition of extra linear layers after the feature extractor of the network and before each incremental task to generate distilled labels. The ICAC approach \cite{li2022incremental} was based on anchored class data centers to promote tighter clustering within each class and better separation between classes. ICAC introduced separable learning to mitigate class imbalance, a learning strategy that computes the loss functions for old and new exemplars separately. MLAKDN \cite{yu2023multilevel} was proposed as a method that combines classification and feature-level knowledge distillation. Ren \etal \cite{ren2024dynamic} introduced a dynamic feature embedding network and a hybrid loss function to optimize the proposed method. Some recent works utilized PTMs as feature extractors. Gao \etal \cite{gao2024sar} introduced a mechanism for enhancing the linear separability of features, utilizing a Resnet-18. Pan \etal \cite{pan2023class} proposed employing a ViT combined with a dynamic query navigation module, which was designed to improve the plasticity of the model. An exemplar-free based method, that does not retain any old-class samples was proposed by Li \etal \cite{li2024sar}, employing a mutual information maximization method to avoid the distribution overlap among classes. A comprehensive summary of the discussed SAR-ATR incremental methods is presented in Table \ref{tab:summary_sar}.

While most studies utilize exemplars, our work introduces an exemplar-free approach based on prototype learning. Furthermore, while previous research has explored the usage of PTMs, we extend the literature by proposing a dual-fusion strategy. This leverages the advantages of combining general features extracted from a PTM with specialized features derived by a custom-designed CNN.

While our focus is on incremental learning for classification in SAR-ATR, it is worth noting that a significant subset of SAR-ATR research is dedicated to detection tasks. These approaches aim to identify multiple objects, such as vehicles and ships, within SAR images, often employing specialized detection frameworks. Although detection-focused methods differ from classification-centric approaches discussed earlier, they contribute valuable insights into handling SAR imagery in real-world scenarios \cite{tang2022cfar, sun2022improved, lv2024efficient, sun2021bifa, li2023sar, yang2022improved, ren2023yolo, guo2022yolox}. Among these, YOLO-based detectors play a significant role in real-world applications of SAR-ATR. A YOLOv3-based detector, called GCN-YOLO, that utilizes a graph convolution network was proposed in \cite{chen2024gcn}, which employs a block attention module to enhance semantic features. Also, a confidence loss was introduced to enable GCN-YOLO efficiency in foreground samples. A SAR-to-optical image translation network was proposed in \cite{lee2023sar}, which employs a modified dense nested U-net to enhance feature translation for target recognition. In addition, a virtual dataset generation method was introduced to improve recognition accuracy by combining 3-D model-based optical images with SAR data. A ship target detection method, named DBW-YOLO, built upon YOLOv7-tiny was introduced in \cite{tang2024dbw} employing  a deformable convolution network for robust feature extraction, coupled with an attention mechanism and an IoU-based loss function to improve detection accuracy.

\begin{figure*} 
\centering
\includegraphics[width=\linewidth]{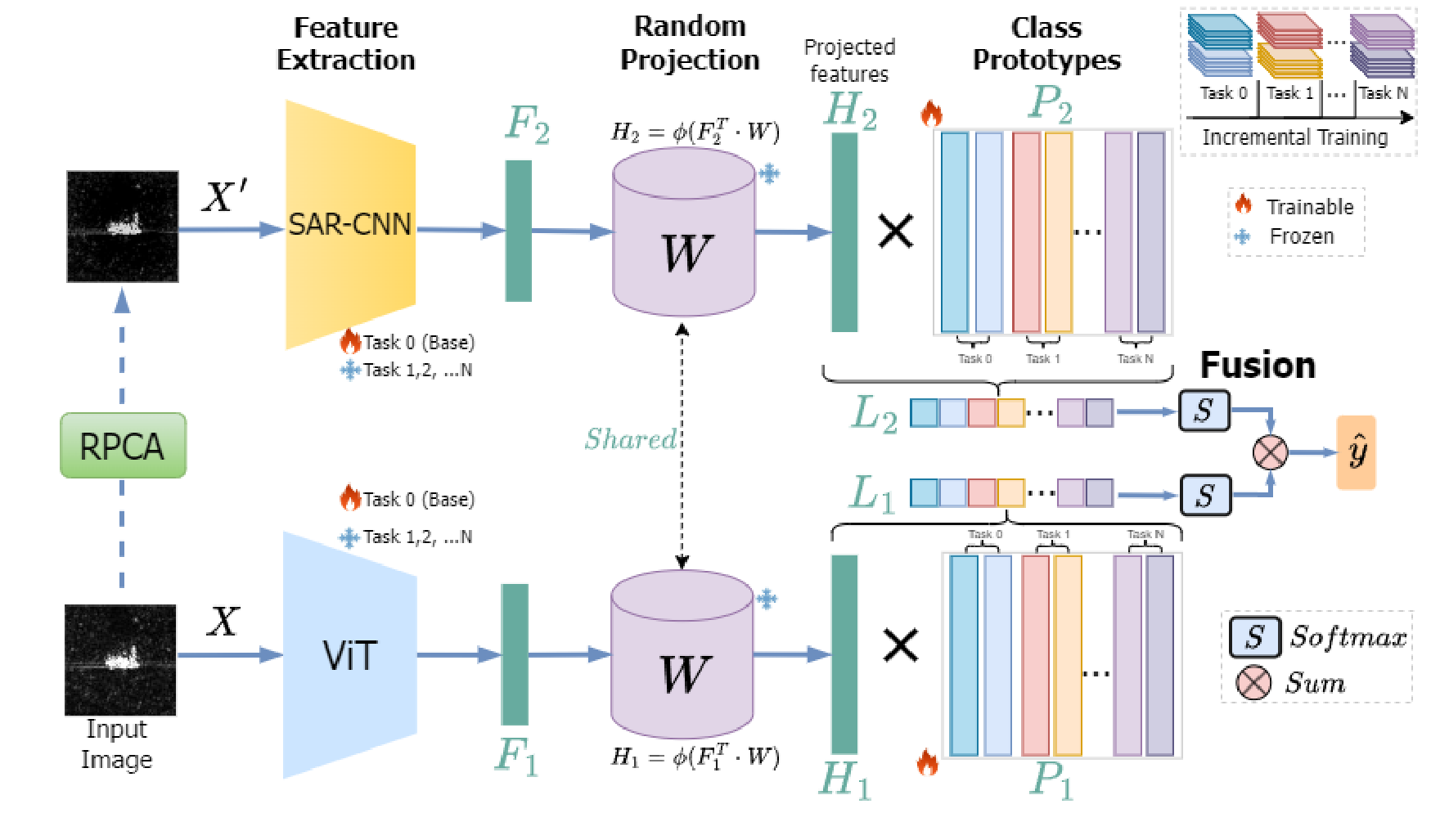}
\caption{Illustration of IncSAR: A late-fusion approach is employed. The input image feeds a ViT  network to extract features $\mathbf{F}_1$. The input image is passed through the filtering RPCA module, and the filtered output feeds the proposed CNN to extract features $\mathbf{F}_2$. The backbone networks are trained only in the base task of CIL, and then their weights are frozen. The extracted features $\mathbf{F}_1$, $\mathbf{F}_2$ are projected into a higher dimensional space using a random projection layer with frozen weights W and an activation function $\phi$, giving $\mathbf{H}_1$, $\mathbf{H}_2$ features respectively. During incremental training, the matrices of the decorrelated class prototypes $\mathbf{P}_1$, $\mathbf{P}_2$ are continually updated for each task. The logits $\mathbf{L}_1$, $\mathbf{L}_2$ are passed to a softmax layer S and an element-wise addition layer to derive the final prediction $\hat{y}$.}
\label{fig:framework}
\end{figure*}

\section{Methodology}
\label{sec: Methodology}

\subsection{Background}
\textbf{CIL}: Unlike the ``traditional'' machine learning setting, where a model is trained on all classes with all training data available at once, in CIL a model sequentially receives new training data with additional classes over time.
More concretely, in a CIL scenario we assume a sequence of $T$ tasks and their corresponding training sets $\mathbf{D}_t$ for $t \in\{1, 2, \ldots, T\}$. A task refers to a set of classes that are disjoint and do not overlap with the classes in other tasks. For each incremental task $t$, the training set is defined as $\mathbf{D}_t= \left\{ \left( x_i, y_i \right) \right\}_{i=1}^{N_t}$, where $N_t$ is the number of training samples in $\mathbf{D}_t$, and $\left( x_i, y_i \right)$ is a training instance with its corresponding label. Here, $y_i \in Y_t $ and  $Y_t$ denotes the label space of task $t$. We refer to $\mathbf{D}_0$ as the base task, and all other tasks as incremental tasks. In typical CIL, it is assumed that there are no overlapping classes between different tasks: $Y_t \cap Y_{t'} = \emptyset \ \forall \ t \neq t'$. During training on task $\mathbf{D}_t$, only data from this task is accessible; data from previous tasks is not available. We adopt an offline learning setting, where we may process the training data multiple times during the current task before moving to the next. After each task, the trained model is evaluated over all seen classes, represented by the set $\mathcal{Y}_t = \bigcup_{i=1}^{t} Y_i $. The aim of CIL is to build a classification model that acquires knowledge of all seen classes $\mathcal{Y}_t$  and performs well not only on the ongoing task but also preserves its performance on previous ones. Particularly, in exemplar-based methods, limited access to old training samples is allowed by storing a small subset of data from previous tasks in a memory buffer, in contrast to exemplar-free methods, which do not retain any previous data.

\subsection{IncSAR framework}
\begin{figure}  
\centering
\includegraphics[width=\linewidth]{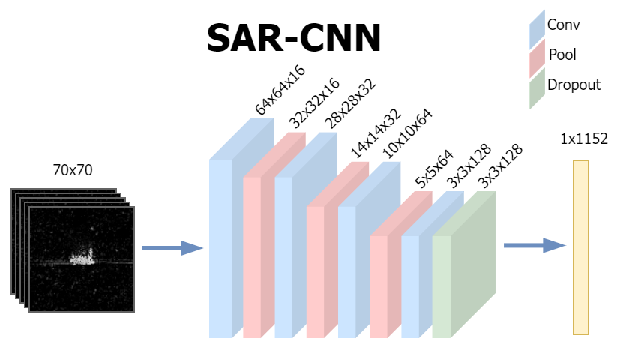}
\caption{Architecture of the proposed SAR-CNN model.}
\label{fig:sar_cnn}
\end{figure}

\begin{figure}  
\centering
\includegraphics[width=\linewidth]{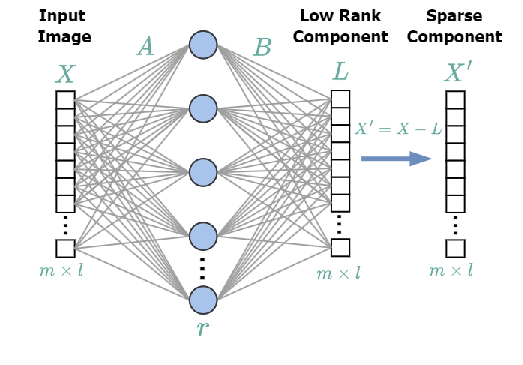}
\caption{Robust PCA procedure resulting in a low-rank and a sparse component.}
\label{fig:RPCA}
\end{figure}

\begin{figure} 
\centering
\includegraphics[width=0.7\linewidth]{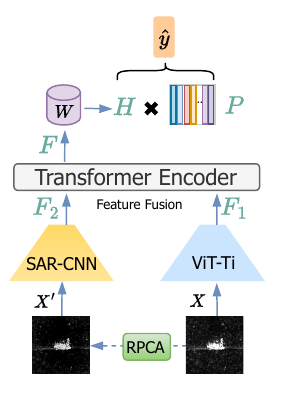}
\caption{Illustration of the proposed feature fusion attention module, demonstrating the integration of features from the ViT-Ti and SAR-CNN branches to produce an enhanced unified representation.}
\label{fig:attention}
\end{figure}

The proposed framework, called IncSAR, is inspired by RanPAC \cite{mcdonnell2024ranpac}, a class-prototype based method that takes advantage of a PTM's feature extraction capabilities. The pipeline of IncSAR is demonstrated in Fig.\ref{fig:framework}. A late-fusion strategy is introduced, comprising two individual branches that take advantage of two different backbones: a pretrained ViT-B/16 model and a custom-designed CNN model, as shown in Fig. \ref{fig:sar_cnn}. The backbone networks are trained individually during the base task, and then the weights are frozen during incremental tasks. A filtering RPCA module is employed before the CNN model, as detailed in Fig. \ref{fig:RPCA}. After the feature extraction, a random projection layer is employed, and the projected features are utilized to compute the class prototypes, while an LDA approach is employed to decorrelate them. Finally, the logits of each branch are integrated to derive the final prediction. Moreover, an IncSAR variant, called IncSAR$_{LAtt}$,  employs an attention mechanism for feature fusion prior to random projection, combining the proposed SAR-CNN model with a pre-trained ViT-Ti \cite{touvron2021training} to further enhance feature extraction and integration, as shown in Fig. \ref{fig:attention}. By leveraging SAR-CNN’s capability to capture spatial and spectral features alongside the transformer's power in capturing global context through self-attention, this hybrid approach enriches the model's representation efficiency. The attention mechanism dynamically prioritizes feature components from SAR-CNN and ViT, effectively balancing fine-grained and high-level semantic information. Moreover, an additional variant is introduced, named IncSAR$_{Lite}$, which follows the base IncSAR pipeline as shown in Fig. \ref{fig:framework}, substituting ViT with TinyViT while maintaining the late fusion strategy. This modification leverages TinyViT’s lightweight architecture, reducing computational demands while preserving essential feature extraction capabilities. 

In a more detailed view, our proposed CNN, denoted as SAR-CNN, constitutes a simple yet effective model. SAR-CNN is trained from scratch in the base task, and then its weights are frozen during the incremental tasks. The input layer takes the RPCA-filtered image $\mathbf{X}'$, with an input size of 70x70 and is followed by a sequence of $4$ convolutional layers, each one followed by a max pooling layer. The activation function for each layer is a ReLU function. The kernel sizes are $7\times7$, $5\times5$, $3\times3$, $3\times3$ and the numbers of kernels are $16$, $32$, $64$, and $128$ respectively. Finally, a dropout layer and a dense layer are applied. The proposed SAR-CNN is depicted in Fig. \ref{fig:sar_cnn}. 

\begin{figure} [ht]
\centering
\begin{subfigure}{0.45\linewidth}
  \centering
  \includegraphics[width= \linewidth]{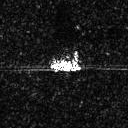}
  \caption{Original image}
  \label{fig:original}
\end{subfigure}
\hfill
\begin{subfigure}{0.45\linewidth}
  \centering
  \includegraphics[width=\linewidth]{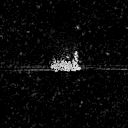}
  \caption{Filtered Image}
  \label{fig:rpca_example}  
\end{subfigure}
\caption{An example of RPCA filtering, employed in MSTAR dataset. Left: original SAR image; right: output of the filtering module.}
\label{fig:comparison}
\end{figure}

The input image $\mathbf{X}$ is filtered by RPCA. The presence of speckle noise in SAR images poses significant challenges, hindering precise analysis and accurate classification. RPCA \cite{rpca} has been utilized in various applications in computer vision. Here, RPCA is utilized as a pre-processing step to denoise SAR images and improve the classification accuracy of the SAR-CNN. An example of RPCA filtering is depicted in Fig. \ref{fig:comparison}.

Let $\mathbf{X}$ be a matrix with a dimension of $m \times l$, representing a noisy SAR image. RPCA defines the problem of decomposing a noise-corrupted data matrix $\mathbf{X} \in \mathbb{R}^{m \times l}$ into two components: a low-rank matrix $\mathbf{L} \in \mathbb{R}^{m \times l}$, which captures the noisy background, and a sparse matrix $\mathbf{X'} \in \mathbb{R}^{m \times l}$, which represents the filtered SAR image. We use an implementation of RPCA as a neural network, as proposed by Han \etal \cite{han2021efficient}, that formulates RPCA as a surrogate optimization problem:

\begin{equation}
\min \|\mathbf{X'}\|_1, \quad \text{subject to} \ \mathbf{X} = \mathbf{L} + \mathbf{X'} \ \text{and} \ \mathbf{L}=\mathbf{A}\mathbf{B}\mathbf{X}
\end{equation}

Here, matrices $\mathbf{A} \in \mathbb{R}^{m \times r}, \mathbf{B} \in \mathbb{R}^{r \times m}$ correspond to the learnable parameters of a network with two linear layers, where $r$ is the desired rank of $\mathbf{L}$. A window size of $128\times128$ is selected for the input image, which is flattened into a one-dimensional vector. The first layer projects the input into the lower-dimensional space of rank $r$, and the second layer maps it back to the original input space giving the low-rank component $\mathbf{L}$. The sparse component $\mathbf{X'}$, which leads to the filtered image, is computed as: 
\begin{equation}
\mathbf{X'} = \mathbf{L}- \mathbf{X}
\end{equation}
The network is trained in the images of the base task and then its weights are frozen during incremental tasks. The entire procedure is depicted in Fig. \ref{fig:RPCA}. 

The pre-trained ViT-B/16 model, initially trained on the ImageNet-21K dataset and fine-tuned on ImageNet-1K, is fine-tuned exclusively on the base task in our study, with its weights frozen during the incremental tasks. The pretrained TinyViT, initially trained on ImageNet-22K\footnote{ImageNet-22K is the same as ImageNet-21K, where the number of classes is 21,841.}, and fine-tuned on ImagNet-1K  employing a fast knowledge distillation method from a CLIP-ViT-L/14, is frozen during the base and incremental tasks. Notably, neither ImageNet-21K nor ImageNet-1K contain SAR images in their training sets. However, despite the absence of SAR data in pre-training, the model demonstrates strong performance when fine-tuned on the base task using the MSTAR dataset. We employ a scale and shift (SSF) method, which was proposed by Lian \etal \cite{lian2022scaling}, to adjust the extracted features to match the distribution of the downstream dataset. This method appends an extra SSF layer after each operation layer of the ViT model. Let $\mathbf{x_{in}}$ be the output of an operation layer with a dimension of $d$. The modulated output $\mathbf{x_{o}}$ is computed by: 
\begin{equation}
\mathbf{x_{o} = \boldsymbol{\gamma} \otimes x_{in} + \boldsymbol{\delta}}
\end{equation}
where  $\otimes$ is an element-wise multiplication operator and $\boldsymbol{\gamma}, \, \boldsymbol{\delta} \in \mathbb{R}^d$ are the scale and shift factors.

During each incremental task, the features $\mathbf{F}$ are extracted individually from each branch. An extra layer, followed by a non-linear function  \textbf{\( \phi \)}, is employed after feature extraction to randomly project the features into a higher-dimensional space $M$. The projected features $\mathbf{H}$ are given by: 
\begin{equation}
\label{eq:proj}
\mathbf{H} = \phi(\mathbf{F}^{\top} \mathbf{W}) 
\end{equation}
This feature transformation is employed to enhance linear separability, and its weights $\mathbf{W}$ are frozen and generated randomly only once before the incremental training. Additionally, a variation of LDA for continual learning is employed to remove correlations between class prototypes. The Gram Matrix $\mathbf{G}$ of features $\mathbf{H}$ is extracted in an iterative manner:   
\begin{equation}
\label{eq:gram}
\mathbf{G} = \sum_{t=1}^{T} \sum_{n=1}^{N_t} \mathbf{H}_{t,n}\otimes \mathbf{H}_{t,n}, 
 \quad 
\end{equation}
The concatenated matrix $\mathbf{C}$ of class prototypes is given by:
\begin{equation}
\label{eq:conc}
\mathbf{C} = \sum_{t=1}^{T} \sum_{n=1}^{N_t} \mathbf{H}_{t,n} \otimes \mathbf{y}_{t,n}
\end{equation}
where $\otimes$ is the outer product, $T$ is the number of incremental tasks and $N_t$ is the number of training samples in each task.
The weights $\mathbf{P}$ represent the decorrelated class prototypes: 
\begin{equation}
\label{eq:deccor}
\mathbf{P = (G + \lambda I)^{-1} C}
\end{equation}
where $\lambda$ is the ridge regression parameter. Parameter $\lambda$ is updated after each task and is optimized by randomly dividing the training data for that task using an 80:20 ratio and selecting the value of $\lambda$ that minimizes the mean square error between targets and the set of predictions. The logits $\mathbf{L}$ are computed by: 
\begin{equation}
\label{eq:logits}
\mathbf{L = H_{\text{test}} P}
\end{equation}
where $\mathbf{H}_{\text{test}}$ refers to the encoded features of a test sample after the random projection layer.

The predictions of each model are integrated to obtain the final decision. A softmax layer $S$ is applied on top of the logits of each model to get the probabilities and an element-wise addition layer to make the final prediction $\hat{y}$:
\begin{equation}
\hat{y} = \arg \max_{c \in \mathcal{Y}_t} \Bigg( S(\mathbf{L}_1^c ) + S(\mathbf{L}_2^c) \Bigg)
\end{equation}
where $\mathbf{L}_1$, $\mathbf{L}_2$ are the logits of SAR-CNN and the logits of ViT respectively, calculated for each class $c$ to select the maximum result for the final prediction.

\textbf{Attention Fusion}. The proposed feature fusion technique leverages a transformer encoder, denoted as $\mathcal{E}$, illustrated in Fig. \ref{fig:attention}. The proposed approach is employed to substitute the late fusion strategy resulting in the IncSAR$_{LAtt}$ variant. Specifically, we utilize ViT-Ti \cite{touvron2021training}, a lightweight version of ViT, and SAR-CNN as feature extractors. The transformer encoder $\mathcal{E}$ consists of $4$ layers with an embedding dimension of $672$, employs $8$ attention heads, and has a feed-forward network dimension of $336$. SAR-CNN is trained from scratch on RPCA-filtered images, while ViT-Ti, which processes the original images, is fine-tuned with the SSF technique. The extracted features from ViT-Ti and SAR-CNN, denoted by $\mathbf{F}_1$ and $\mathbf{F}_2$, respectively, are concatenated and subsequently input into $\mathcal{E}$ for further processing:

\begin{equation}
\mathbf{F} = \mathcal{E}([\mathbf{F}_1; \mathbf{F}_2])
\end{equation}

ViT-Ti, SAR-CNN and $\mathcal{E}$ are trained in an end-to-end way in the base task. 
The extracted features $\mathbf{F}$ are then randomly projected, and prototypes $\mathbf{P}$ and logits $\mathbf{L}$  are calculated with the same methodology, as described in Eqs.~(\ref{eq:proj})--(\ref{eq:logits}). To create a more lightweight framework and reduce the network's computational cost and training time, we calculate the parameter $\lambda$ during the base task and keep it constant throughout the incremental tasks. The final prediction $\hat{y}$ is computed as:

\begin{equation}
\hat{y} = \arg \max_{c \in \mathcal{Y}_t} ( \mathbf{L}^c )
\end{equation}

\section{Experiments}
\label{sec:exp}

\subsection{Datasets and Experimental settings}
\label{sssec:expsettings}

\begin{figure*}[ht]
    \centering
    \includegraphics[width=\textwidth]{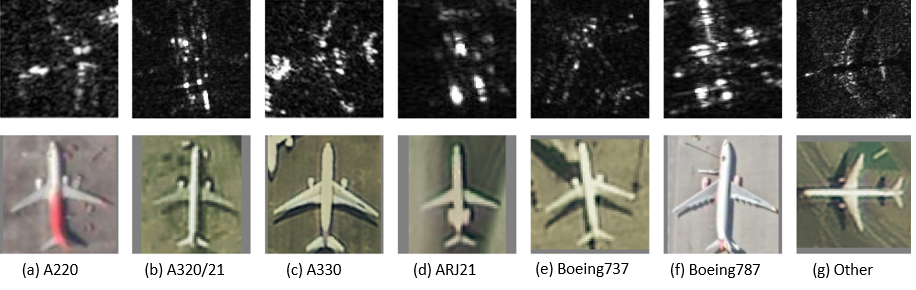}
    \caption{Example aircraft photos (bottom row) and corresponding SAR images (top row) for seven different aircraft classes of SAR-AIRcraft-1.0 dataset: A220, A320/321, A330, ARJ21, Boeing 737, Boeing 787, and 'other.' The images illustrate the visual and radar characteristics of each class. The figure is adapted from \cite{zhirui2023sar}.}
    \label{fig:datasetAIR}
\end{figure*}

\textbf{Datasets.}
To evaluate IncSAR for classifying SAR images, we initially employ the MSTAR dataset \cite{ross1998standard}. MSTAR is a publicly available benchmark dataset of SAR images that contains $10$ ground mobile targets, as shown in Table \ref{tab:mstar}. SAR images are acquired at two different angles of depression, i.e., 15° and 17°. Images at 17° are used for training, and images at 15° are used for testing. The OpenSARShip \cite{huang2017opensarship} dataset is also employed in the conducted experiments for generalization purposes, as done in \cite{tang2022incremental}. OpenSARShip contains $11,346$ SAR ship chips, integrated with automatic identification system (AIS) messages. The dataset covers 17 AIS types collected from 41 Sentinel-1 SAR images. Three ship types are selected, i.e., bulk carrier, container ship, and tanker, under the VV polarization mode. We randomly select $300$ samples from each class and split them into training and test sets with an 80:20 ratio. Additionally, the SAR-AIRcraft-1.0 dataset \cite{zhirui2023sar} is used for further experiments. This dataset provides images in four different sizes: $800$ × $800$, $1000$ × $1000$, $1200$ × $1200$, and $1500$ × $1500$ pixels, featuring $16,463$ aircraft instances across seven categories: A220, A320/321, A330, ARJ21, Boeing 737, Boeing 787, and an 'other' category. It is characterized by complex scenes, rich categories, dense targets, noise interference, and multi-scale data, making it particularly suited for various SAR-based tasks. The SAR-AIRcraft-1.0 configuration is presented in Table \ref{tab:AIR}. Figure \ref{fig:datasetAIR} illustrates representative samples from the SAR-AIRcraft-1.0 dataset. These three datasets were selected due to their prevalent use in related literature, as can be seen in Table \ref{tab:summary_sar}.

\begin{table}[h]
    \caption{Configuration of MSTAR dataset.}
    \label{tab:mstar}
    \centering
    \resizebox{\columnwidth}{!}{%
        \begin{tabular}{cl|cccc}
            \hline
            \multirow{2}{*}{Class} & \multirow{2}{*}{Class name} & \multicolumn{2}{c}{Training set} & \multicolumn{2}{c}{Testing set} \\
             & & Depression & Number & Depression & Number \\
            \hline
            \hline
            $0$ & BTR60 & $17^{\circ}$ & $256$ & $15^{\circ}$ & $195$ \\
            $1$ & T72 & $17^{\circ}$ & $232$ & $15^{\circ}$ & $196$ \\
            $2$ & 2S1 & $17^{\circ}$ & $299$ & $15^{\circ}$ & $274$ \\
            $3$ & T62 & $17^{\circ}$ & $299$ & $15^{\circ}$ & $273$ \\
            $4$ & ZIL131 & $17^{\circ}$ & $299$ & $15^{\circ}$ & $274$ \\
            $5$ & ZSU234 & $17^{\circ}$ & $299$ & $15^{\circ}$ & $274$ \\
            $6$ & BRDM2 & $17^{\circ}$ & $298$ & $15^{\circ}$ & $274$ \\
            $7$ & D7 & $17^{\circ}$ & $299$ & $15^{\circ}$ & $274$ \\
            $8$ & BMP2 & $17^{\circ}$ & $233$ & $15^{\circ}$ & $195$ \\
            $9$ & BTR70 & $17^{\circ}$ & $233$ & $15^{\circ}$ & $196$ \\
            \hline
        \end{tabular}%
    }
\end{table}

\begin{table}[ht]
\centering
\caption{Configuration of SAR-AIRcraft-1.0 dataset.}
\begin{tabular}{cl|cc}
\hline
Class  & Class name & Training Set & Testing set \\ \hline \hline
   $0$& Other      &$2000$&$200$\\ 
       $1$& A220       &$2000$&$200$\\  
       $2$& Boeing787  &$2000$&$200$\\  
       $3$& Boeing737  &$2000$&$200$\\ \hline
 $4$& A320  &$1571$&$200$\\  
       $5$& ARJ21  &$987$ &$200$\\  
       $6$& A330   &$209$ &$200$\\ \hline
\end{tabular}
\label{tab:AIR}
\end{table}

\textbf{Evaluation Protocol.}
A suite of evaluation metrics is employed to assess the performance of IncSAR. Top-1 accuracy in the $t^{th}$ task is denoted as $A_t$. The accuracy in the last incremental task, denoted as $A_L$, is a suitable metric to measure the overall accuracy among all classes. The average incremental accuracy $\bar{A}$ takes into consideration the overall accuracy scores along all incremental tasks: $\bar{A}=\frac{1}{T}\sum_{t=0}^{T} A_t $. Also, we utilize the performance dropping rate PD $ = A_0 - A_L$ and the performance dropping rate per task PD$_t= A_0 - A_t$, where $A_0$ denotes accuracy in the base task and $A_t$ accuracy in the $t^{th}$ incremental task. PD is an established metric in the literature, that tries to quantify how much forgetting takes place in the overall procedure.

\textbf{Training Details.}
Experiments are implemented using the PyTorch \cite{paszke2019pytorch} framework and PILOT \cite{sun2023pilot}, a pre-trained model-based continual learning toolbox. Two different data augmentation approaches are employed for each backbone. For ViT-B/16, the original images are simply padded to a size of $224\times224$. For SAR-CNN, training images are filtered by RPCA followed by common transformations, such as cropping to $32\times32$, resizing to $70\times70$, and random horizontal flipping. Targets in the MSTAR and OpenSARShip datasets are centered in the middle of the image, allowing cropping to discard unwanted noise peripheral to the target. The SAR-CNN branch is trained for $30$ epochs and the ViT/B-16 branch for $10$ epochs, both using a learning rate of $0.01$, a weight decay of $0.0005$, and stochastic gradient descent optimizer with a momentum of $0.9$. The dimension of random projection is set to $M=10000$.  In IncSAR$_{LAtt}$, the network is trained for $15$ epochs with the same hyperparameters.

\begin{table}[t]
\caption{Comparison with prior works across each incremental task on MSTAR dataset. Base incremental task consists of 4 classes, and each incremental task of 1 class.}
\centering
\renewcommand{\arraystretch}{1.2}
\resizebox{\columnwidth}{!}{
\begin{tabular}{lcccccccccc}  
\hline
\multirow{2}{*}{Method} & \multicolumn{7}{c}{Accuracy in each task (\%)} & \multirow{2}{*}{PD $\downarrow$} & \multirow{2}{*}{$\bar{A}$ $\uparrow$} \\
\cline{2-8}
& $0$ & $1$ & $2$ & $3$ & $4$ & $5$ & $6$ & & \\
\hline
DualPrompt \cite{yang2023dual} & $85.50$ & $66.17$ & $53.97$ & $45.57$ & $39.43$  & $36.03$ & $33.73$ & $51.77$  & $51.48$ \\
iCaRL \cite{rebuffi2017icarl} & $70.90$ & $72.85$ & $73.49$ & $76.48$ & $58.95$  & $55.94$ & $52.66$ & $18.24$  & $65.89$ \\
FOSTER \cite{wang2022foster} & $63.54$ & $84.90$ & $71.27$ & $69.72$ & $67.65$ & $61.01$ & $59.42$ & ${4.12}$ & $68.21$ \\
SimpleCIL \cite{zhou2023revisiting} & $88.81$ & $88.61$ & $86.88$ & $86.31$ & $84.46$ & $80.48$ & $76.87$ & $84.63$ & $11.94$ \\
aper\_adapter \cite{zhou2023revisiting} & $89.13$ & $88.70$ & $87.01$ & $86.65$ & $84.71$ & $80.62$ & $76.99$ & $12.14$ & $84.83$ \\
aper\_ssf \cite{zhou2023revisiting} & $93.07$ & $94.55$ & $92.40$ & $91.02$ & $90.02$ & $85.64$ & $80.95$ & $12.12$ & $89.66$ \\
MEMO \cite{zhou2022model} & $91.36$ & $93.40$ & $93.61$ & $92.90$ & $91.64$ & $88.83$ & $85.15$ & $6.21$ & $90.98$ \\
FeCAM \cite{goswami2024fecam} & $94.56$ & $94.64$ & $94.55$ & $94.72$ & $93.81$ & $91.16$ & $87.59$ & $6.97$ & $93.00$ \\
RanPAC \cite{mcdonnell2024ranpac} & ${98.61}$ & ${98.68}$ & ${98.12}$ & ${98.35}$ & ${97.49}$ & ${94.93}$ & ${94.23}$ & $4.38$ & ${97.20}$ \\
\hline 
Pan et. al \cite{pan2023class} & $98.49$ & - & - & - & - & - & $74.65$ & - & - \\
ICAC \cite{li2022incremental} & $\underline{99.49}$ & $98.04$ & $96.76$ & $95.65$ & $94.83$ & $93.42$ & $91.76$ & $4.66$ & $96.65$ \\
\hline
IncSAR  & $\mathbf{100.00}$ & $\mathbf{99.75}$ & $\underline{99.39}$ & $\underline{99.43}$ & $\underline{99.41}$ & $97.71$ & $\mathbf{99.22}$ & $\mathbf{0.78}$ & $\underline{99.27}$ \\
IncSAR$_{Lite}$  & $99.25$ & $98.60$ & $98.92$ & $98.81$ & $98.23$ & $\mathbf{98.83}$ & $97.77$ & $1.48$  & $98.63$ \\
IncSAR$_{LAtt}$  & $99.47$ & $\underline{99.59}$ & $\mathbf{99.66}$ & $\mathbf{99.83}$ & $\mathbf{99.75}$ & $\underline{98.70}$ & $\underline{98.39}$ & $\underline{1.08}$ & $\mathbf{99.34}$ \\
\hline
\end{tabular}
}
\label{tab:MSTAR_B4Inc1}
\end{table}

\subsection{Competing methods}
\label{sssec:SOTA}
The proposed IncSAR framework is compared with state-of-the-art incremental learning methods that use PTMs, as well as with state-of-the-art CIL algorithms designed specifically for SAR-ATR recognition.  Moreover, two variants of IncSAR are tested in the experiments on the MSTAR dataset: one incorporates ViT-Ti with the attention module, and the other uses TinyViT with the late-fusion strategy. Two main incremental setups consistent with the literature were employed for the evaluation of the proposed framework. 

In the first setup, denoted as B4Inc1, the base task comprises 4 classes, while each incremental task consists of a single class. The class order is shown in Table \ref{tab:mstar}, following the same order as in \cite{pan2023class}.  Nine CIL state-of-the-art methods were employed together with two state-of-the-art methods from the field of SAR-ATR, namely, DualPrompt, iCaRL, FOSTER, aper\_adapter, aper\_ssf, SimpleCIL, MEMO, FeCAM, RanPAC, ICAC, and a method proposed by Pan et al. \cite{pan2023class}. The PILOT \cite{sun2023pilot} toolbox is used to test the state-of-the-art methods in a standardized manner. The proposed IncSAR achieves an average accuracy of $99.27\%$, demonstrating very strong performance in classifying SAR images, and outperforming the state-of-the-art RanPAC method, which yields an accuracy  of $97.2$\%. IncSAR also surpasses the state-of-the-art ICAC approach by $7.52\%$ in terms of $A_L$ and by $2.64\%$ in terms of $\bar{A}$. IncSAR demonstrates a relative improvement of $81.07\%$ regarding performance drop, attaining $0.78\%$ and outperforming FOSTER, which yields a performance drop of $4.12\%$. ICAC is lagging behind IncSAR and FOSTER with a performance drop of $4.66\%$. IncSAR$_{Lite}$  variant derives an $\bar{A}$ of $98.63\%$, outperforming state-of-the-art methods, while IncSAR$_{LAtt}$ reaches an average accuracy of $99.34\%$ making it the top-performing method on the MSTAR dataset. It also achieves a performance drop of $1.08\%$ making the second best result yielding an improvement of $73.79\%$ compared to state-of-the-art methods. The results are detailed in Table \ref{tab:MSTAR_B4Inc1}.

\begin{table}[htb]
    \caption{Comparison with state-of-the-art in each incremental task on the MSTAR dataset. The classes are equally divided into five tasks, each consisting of two classes.}
    \label{tab:MSTAR_B2Inc2}
    \centering
    \resizebox{\columnwidth}{!}{%
        \begin{tabular}{lccccccc}  
            \hline
            \multirow{2}{*}{Method} & \multicolumn{5}{c}{Accuracy in each task (\%)} & \multirow{2}{*}{PD $\downarrow$} & \multirow{2}{*}{$\bar{A}$ $\uparrow$} \\
            \cline{2-6}
            & $0$ & $1$ & $2$ & $3$ & $4$  \\
            \hline
            Hu \etal \cite{hu2022incremental} & $99.60$ & $87.96$ & $84.60$ & $83.89$ & $84.60$ & $15.00$ & $88.13$ \\
            SSF-IL \cite{gao2024sar} & - & - & - & - & - & - & $98.05$ \\
            MLAKDN \cite{yu2023multilevel} & $99.64$ & $99.82$ & $\underline{98.98}$ & $96.87$ & $94.50$ & $5.14$ & $97.96$ \\
            DERDN \cite{ren2024dynamic} & $99.63$ & $99.05$ & $97.71$ & $95.48$ & $93.70$ & $5.93$ & $97.11$ \\
            HPecIL \cite{tang2022incremental} & $99.45$ & $98.83$ & $98.79$ & $96.70$ & $96.16$ & $3.29$ & $97.99$ \\
            Zhou \etal \cite{zhou2022sar} & - & - & - & - & - & - & $97.73$ \\
            RanPAC \cite{mcdonnell2024ranpac} & $98.18$ & $98.51$ & $96.45$ & $95.15$ & $95.13$ & $3.05$ & $96.68$ \\
            \hline
            IncSAR & $100.00$ & $\underline{99.89}$ & $98.94$ & $\underline{99.15}$ & $\underline{99.22}$ & $\underline{0.78}$ & $\underline{99.44}$ \\
            IncSAR$_{Lite}$ & $\mathbf{100.00}$ & $\mathbf{100.00}$ & $\mathbf{99.43}$ & $\mathbf{99.73}$ & $\mathbf{99.38}$ & $\mathbf{0.62}$ & $\mathbf{99.70}$ \\
            IncSAR$_{LAtt}$ & $100.00$ & $\underline{99.89}$ & $97.94$ & $97.60$ & $97.90$ & $2.10$ & $98.66$ \\
            \hline
        \end{tabular}
    }
\end{table}

\begin{figure} 
\centering
\includegraphics[width=\columnwidth]{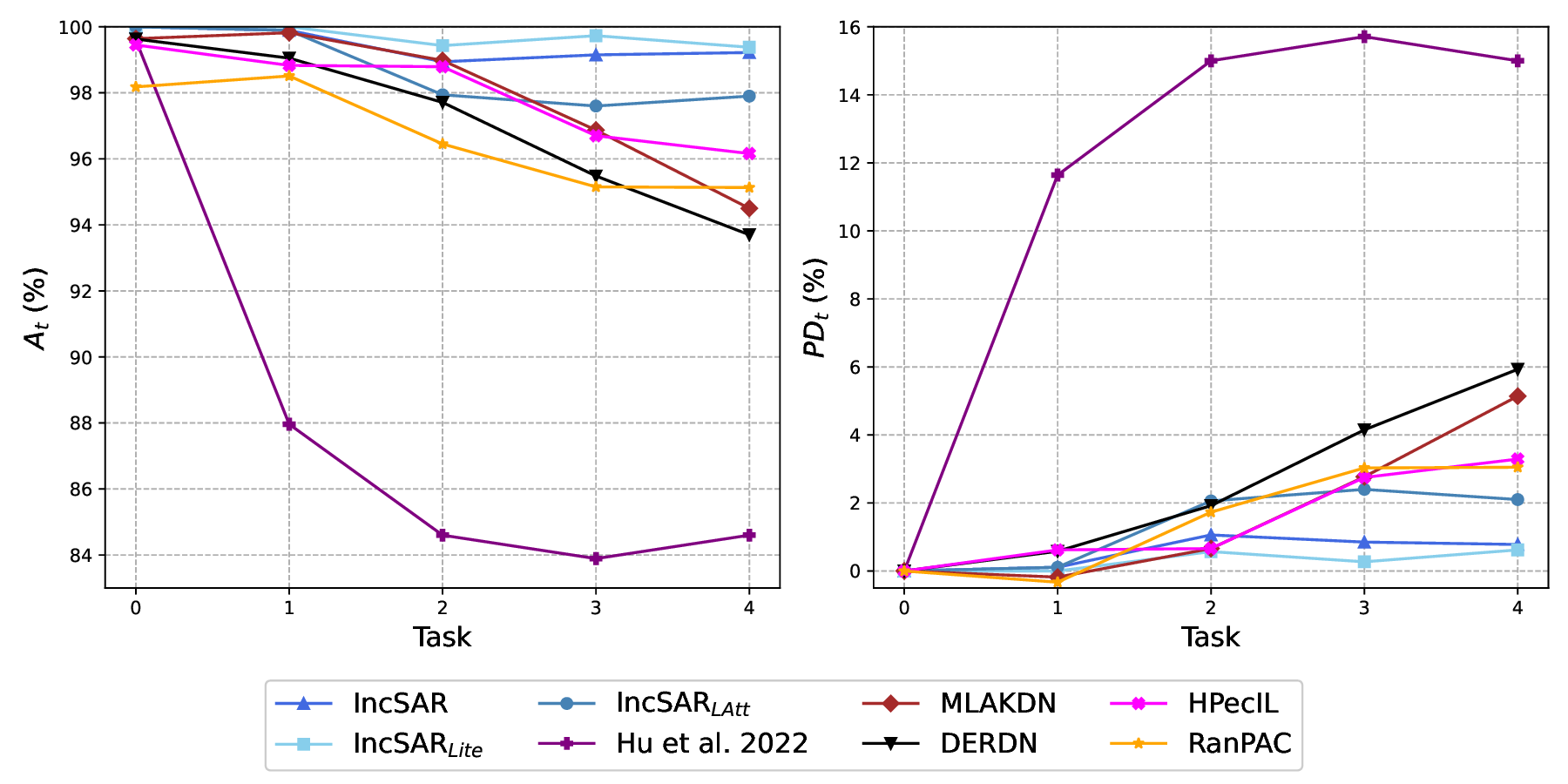}
\caption{Comparison with state-of-the-art methods on MSTAR dataset. Classification accuracy $A_t$ and performance drop PD$_t$ of each incremental task $t$ are depicted.}
\label{fig:b2inc2}
\end{figure}

In the second setup, denoted as B2Inc2, all incremental tasks are evenly split in two classes. The same class order is employed, as in \cite{tang2022incremental,ren2024dynamic,yu2023multilevel}.  The vast majority of methods in the base task demonstrate accurate results achieving over $99\%$. As tasks increase sequentially, catastrophic forgetting occurs, leading to performance drops, as shown in Fig. \ref{fig:b2inc2}. However, IncSAR exhibits robust performance over all incremental tasks, showcasing the lowest PD compared to the state-of-the-art. Experimental results attest to the remarkable ability of IncSAR to resist catastrophic forgetting achieving a PD of $0.78$\%, outperforming RanPAC, which attains a PD of $3.05\%$, demonstrating a relative improvement of $74.43\%$. IncSAR surpasses MLAKDN by $5$\% in $A_L$ and by $84.83$\% in PD. IncSAR yield an improvement of $3.18$\% in $A_L$ and  $76.07$\% in PD, when compared to HPecIL. IncSAR outperforms its state-of-the-art competitors resulting in an average accuracy of $99.44\%$.
Moreover, noteworthy improvements are observed when the IncSAR$_{Lite}$ variant of IncSAR is employed along with the late fusion module, yielding a performance drop of $0.62\%$. 
Similar results are recorded with respect to average accuracy and last task accuracy achieving a $99.7\%$ and $99.38\%$, respectively, surpassing in both cases the state-of-the-art approaches. It should be also noted that IncSAR$_{Lite}$ achieves the best results in all incremental tasks in MSTAR  dataset. The third variant of IncSAR, which incorporates the attention module, delivers exceptional performance, surpassing state-of-the-art methods in both average accuracy and minimizing performance drop.
Note that IncSAR, and its variants, does not use exemplars, unlike HPecIL, MLAKDN, and DERDN. This makes it an even more challenging scenario, as it lacks direct access to past data, unlike exemplar-based methods, which preserve and replay stored samples to mitigate catastrophic forgetting. The results are shown in Table \ref{tab:MSTAR_B2Inc2}.

\begin{table}[t]
    \caption{Comparative analysis of different variations of IncSAR framework on the MSTAR dataset.}
    \label{tab:param}
    \centering
    \resizebox{\columnwidth}{!}{%
        \begin{tabular}{lccc}
             
            Method & Parameters (M) & MACs (G) & Training Time (s)   \\
            \hline
            IncSAR & $106$ & $17.62$ & $466$  \\
            IncSAR$_{Lite}$ & $21$ & $1.34$ & $359$  \\
            IncSAR$_{LAtt}$ & $17$ & $1.30$ & $92$ \\
            \hline
        \end{tabular}%
    } 
\end{table}

A comparative analysis of the variations of IncSAR on the MSTAR dataset using the first setup is shown in Table \ref{tab:param}. All experiments were conducted on an RTX 4090 GPU with Multiply-Accumulate Operations (MACs) calculated using the fvcore \cite{fvcore} library. IncSAR requires the most computational resources, consisting of $106$M parameters and $17.62$G MACs. In contrast, IncSAR$_{Lite}$ has $80.18$\% fewer parameters and achieves a 22.9\% reduction in training time compared to IncSAR. IncSAR$_{LAtt}$ maintains similar computational requirements to IncSAR$_{Lite}$ in terms of MACs, but has $19\%$ fewer parameters and exhibits a notable reduction of $74.37$\% in training time. These results demonstrate that the IncSAR framework can be effectively utilized in resource-constrained scenarios without compromising performance.

\begin{table*}[h]
    \caption{Results in cross-dataset testing. Three classes of the OpenSarShip dataset are added in the last incremental tasks to evaluate the generalization ability of IncSAR.}
    \label{tab:mstar_opensar}
    \centering
    \resizebox{2\columnwidth}{!}{%
        \begin{tabular}{l*{7}{c}cc}  
            \hline
            & \multicolumn{7}{c}{Accuracy in each task (\%)} & \multirow{2}{*}{PD $\downarrow$} &  \multirow{2}{*}{$\bar{A}$ $\uparrow$} \\
            \cline{2-8}
            \multirow{2}{*}{Method} & ZIL131/D7 & BTR70/T72 & BMP2/BRDM2 & T62/BTR60 & 2S1/ZSU234 & Bulk Carrier/Container & Tanker &  \\
            & $0$ & $1$ & $2$ & $3$ & $4$ & $5$ & $6$ &  \\
            \hline
            iCaRL \cite{rebuffi2017icarl} & $99.27$ & $99.25$ & $93.40$ & $92.32$ & $93.88$ & $90.62$ & $89.99$ & $9.28$ & $94.1$ \\
            ECIL \cite{tang2022incremental} & $99.45$ & $98.82$ & $98.54$ & $95.31$ & $93.65$ & $94.20$ & $89.34$ & $10.11$ & $95.61$ \\
            ECIL+ \cite{tang2022incremental} & $99.63$ & $98.51$ & $98.08$ & $96.69$ & $96.41$ & $94.43$ & $92.26$ & $7.37$ & $96.57$ \\
            HPecIL \cite{tang2022incremental} & $99.45$ & $98.83$ & $98.79$ & $96.70$ & $96.16$ & $95.89$ & $94.07$ & $5.38$ & $97.10$ \\
            \hline
            IncSAR & $\mathbf{100.00}$ & $\mathbf{99.89}$ & $\underline{98.94}$ & $\underline{99.15}$ & $\underline{99.22}$ & $97.13$ & $96.01$ & $3.99$ & $\underline{98.62}$ \\
            IncSAR$_{Lite}$ & $\underline{99.82}$ & $99.89$ & $\mathbf{99.65}$ & $\mathbf{99.57}$ & $\mathbf{99.59}$ & $\mathbf{98.27}$ & $\underline{96.08}$ & $\underline{3.74}$ & $\mathbf{98.98}$ \\
            IncSAR$_{LAtt}$ & $100.00$ & $99.89$ & $97.94$ & $97.60$ & $97.90$ & $\underline{97.25}$ & $\mathbf{96.66}$ & $\mathbf{3.34}$ & $98.17$ \\
            \hline
        \end{tabular}
    }
\end{table*}

\subsection{Evaluation of generalization ability}
\label{sssec:opensar}
\begin{figure} 
\centering
\includegraphics[width=\columnwidth]{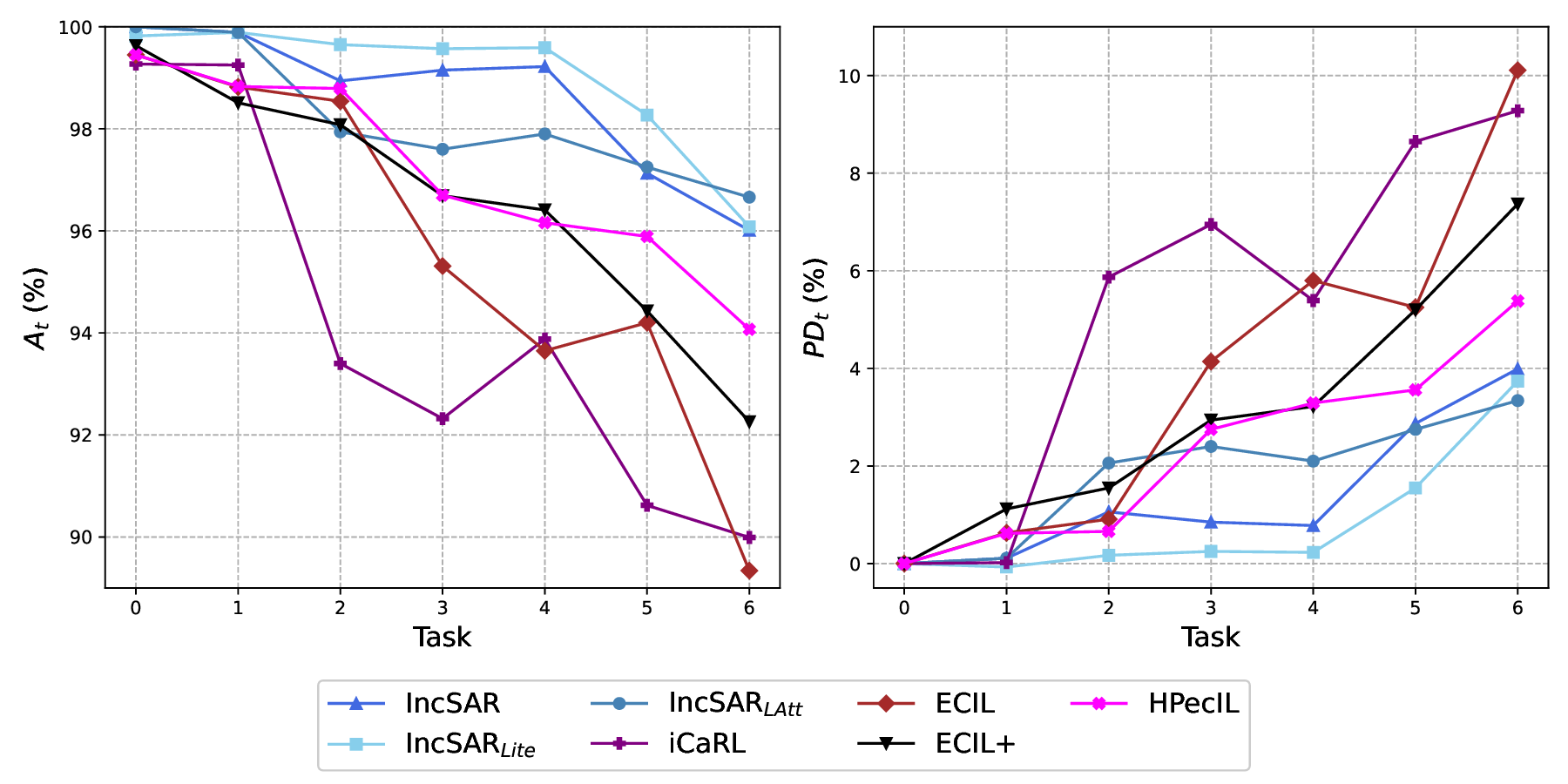}
\caption{Comparison with state-of-the-art methods for testing the generalization ability of the proposed framework.}
\label{fig:mstar_opensar}
\end{figure}

For the evaluation of the generalization ability of IncSAR and its variants, three classes from OpenSARShip are added in the last incremental tasks as done in \cite{tang2022incremental}, for fair comparison. The setup and experimental results are listed in Table \ref{tab:mstar_opensar}. The accuracy in each task and the performance drop for various state-of-the-art methods are depicted in Fig. \ref{fig:mstar_opensar}.
Despite the different distribution and varying sizes of targets in the OpenSARShip dataset, IncSAR outperforms its competitors, attaining an average accuracy of $98.62\%$, while HPecIL is lagging behind with an accuracy of $97.1\%$.  IncSAR is also the top performing approach in the last incremental task, demonstrating an accuracy of $96.01\%$, while HPecIL and ECIL+ yielded $94.07\%$ and $92.26\%$, respectively. The proposed IncSAR demonstrates superior results in all incremental tasks compared to state-of-the-art methods and the iCaRL one, which acts as a baseline. It is worth mentioning that IncSAR derives a value of $3.99$ in performance drop, significantly outperforming HPecIL, which attains a value of $5.38$. This indicates that IncSAR maintains high accuracy across all incremental tasks, effectively addressing the challenge of catastrophic forgetting. This is particularly significant in demanding generalization experiments that closely mirror real-world applications. The IncSAR$_{Lite}$ variant also achieved an average accuracy of $98.98\%$ outperforming all its  competitors. Moreover, its robust performance is attested by the performance drop rate of $3.74\%$. The IncSAR$_{LAtt}$ variant derives the top performance drop value of $3.34\%$, while it reaches a $98.17\%$ in terms of average accuracy. The rest of the methods demonstrate higher values reaching a PD of $10.11$ for the ECIL method. Compared to HPecIL, IncSAR improves by $2.06$\% in $A_L$ and by $25.84$\% in PD. These results attest to the remarkable efficacy of IncSAR and its variants in handling the cross-dataset challenges posed by the OpenSARShip dataset. 

\subsection{Cross-domain evaluation}

To further assess the robustness of the IncSAR framework, we conducted a cross-domain evaluation  that challenges the model's ability to generalize across distinct SAR image datasets, namely, SAR-AIRcraft-1.0, MSTAR, and OpenSARShip. The goal of this setup is to simulate a realistic and demanding scenario where the model first learns to classify aircraft images and must incrementally adapt to recognize military vehicles and ships, all with minimal forgetting of previously learned classes. IncSAR is trained on four base classes, as shown in Table \ref{tab:AIR}, according to B4Inc1 scenario.  In each incremental task, one additional class is introduced from the SAR-AIRcraft-1.0 dataset until all aircraft classes are learned. Afterwards, the model transitions to learning 10 additional classes of military vehicles from the MSTAR dataset, followed by the three classes of OpenSARShip, where it encounters a new set of ship images.  This progressive training, moving from aircraft to military vehicles and finally to ships, simulates a cross-domain learning path requiring the model to handle increasingly diverse visual categories without compromising prior knowledge.

\begin{figure} 
\centering
\includegraphics[width=\columnwidth]{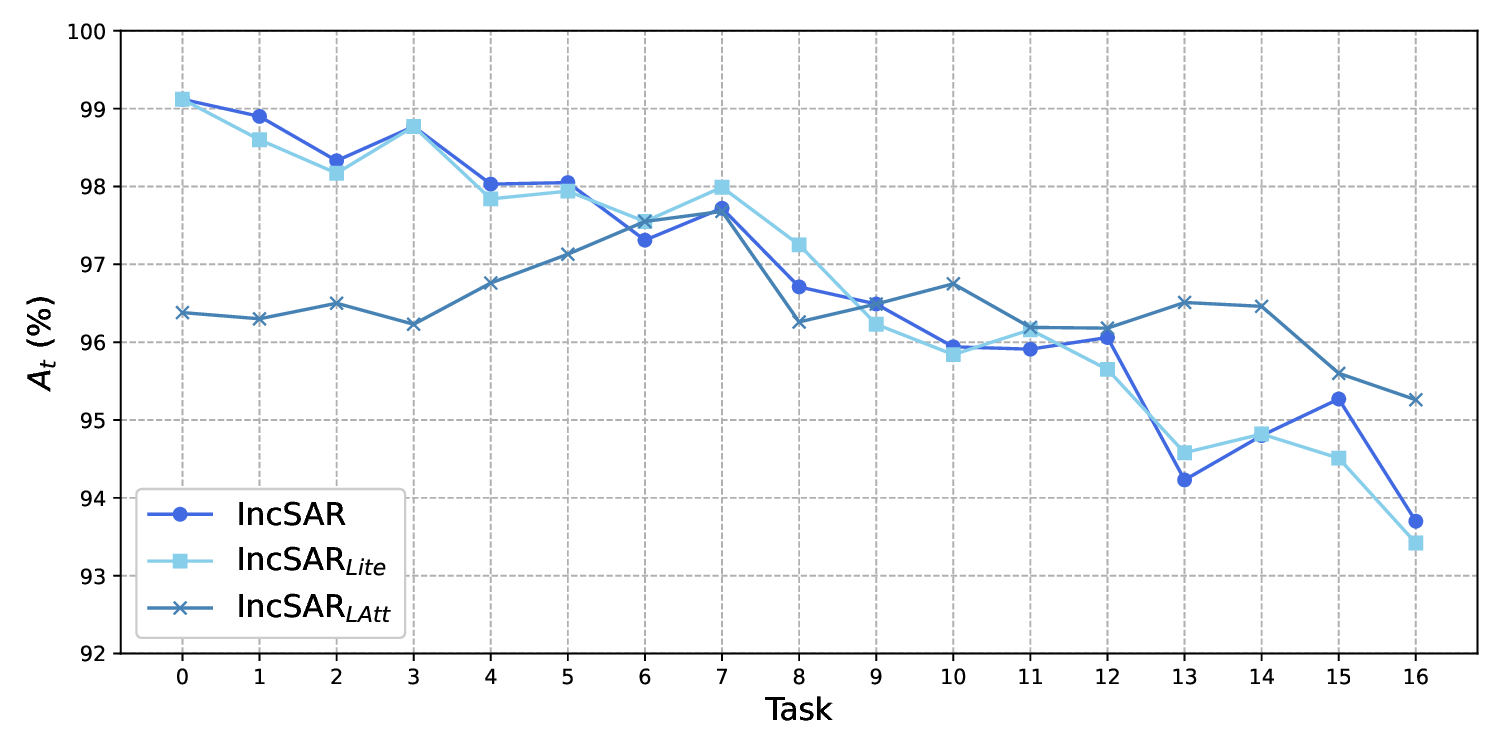}
\caption{Cross-domain evaluation combining SAR-AIRcraft-1.0, MSTAR, and OpenSARShip.}
\label{fig:aircraft_mstar_opensar}
\end{figure}

IncSAR achieves a high $\bar{A}$ of $96.78\%$ and an accuracy of $93.7\%$ on the last task, showing that it generalizes effectively across the three domains. However, the model exhibits a PD of $5.42\%$, indicating some degree of forgetting as new classes and domains are introduced. This result suggests that while IncSAR can manage a cross-domain shift, the transition between disparate categories introduces challenges for knowledge retention. IncSAR$_{Lite}$ yields similar outcomes, achieving an $\bar{A}$  of $96.73\%$ with a slightly higher PD of $5.7\%$. This variant performs comparably to IncSAR but demonstrates a marginally larger performance drop. IncSAR$_{LAtt}$ demonstrates a distinct advantage in terms of PD, achieving the lowest performance drop at $3.08\%$. It also reports a higher last-task accuracy $A_L = 94.67\%$, highlighting the attention module's utility in mitigating forgetting and retaining learned features when adapting to new domains. However, the average accuracy $\bar{A} = 96.57\%$ is slightly lower than the other two variants. The cross-domain results are shown in Table \ref{tab:crossdataset} and in a more detailed view the accuracy in each task is depicted in Fig. \ref{fig:aircraft_mstar_opensar}.

These results collectively demonstrate the capability of the IncSAR framework in tackling cross-domain SAR classification tasks. IncSAR$_{LAtt}$’s superior performance in mitigating forgetting highlights the effectiveness of attention mechanisms, especially in complex cross-domain scenarios. However, all variants show some performance drop, indicating that cross-domain incremental learning remains a challenging task, particularly when the target categories vary greatly in visual characteristics and domain-specific features.

\begin{table}[htb]
    \caption{Cross-domain evaluation combining SAR-AIRcraft-1.0, MSTAR, and OpenSARShip.}
    \label{tab:crossdataset}
    \centering
        \begin{tabular}{l ccc}
             
            Method &  $\bar{A}$ & PD & $A_L$\\
            \hline
             IncSAR& ${96.78}$ &  ${5.42}$ & ${93.7}$ \\
             IncSAR$_{Lite}$ & ${96.73}$ & ${5.70}$ & ${93.42}$  \\
             IncSAR$_{LAtt}$ & ${96.57}$ & ${3.08}$ & ${94.67}$  \\
            \hline
        \end{tabular}%
\end{table}

\newpage
\subsection{Ablation studies}
\label{sssec:ablationj}
\subsubsection{Contribution of IncSAR modules}
The proposed IncSAR framework benefits from multiple modules, including RPCA, SSF adaptation of ViT, fusion of the individual SAR-CNN and ViT branches, RP, and LDA. To explore the contribution of these modules, a series of experiments were conducted and the results are shown in Table \ref{tab:ablationPretrained}. The ablation experiments were conducted on the MSTAR dataset using the B4Inc1 setup. First, we assess the performance of ViT-B/16 as a network backbone, where the models' weights remain frozen throughout the training process. This serves as a baseline to understand the capabilities of the pre-trained ViT-B/16 model without any fine-tuning and other components enabled, achieving $76.87\%$ in terms of $A_L$. When ViT-B/16 is adapted with the SSF technique, RP, and LDA the model improves $A_L$ by $27.09\%$, showing that if there is sufficient data in the base task, adapting the PTM to the downstream dataset can be effective. Moreover, experiments employing only the single branch with SAR-CNN were conducted achieving an average accuracy of $96.45\%$. We demonstrate the improvement achieved by RPCA filtering, when using the proposed SAR-CNN architecture as a backbone, where IncSAR attains an average accuracy of $98.58\%$ compared to the resulting accuracy without employing the RPCA module. This indicates that RPCA enhances SAR-CNN's ability to provide more distinguishable features, leading to better class separability. When both backbone branches are combined, the late-fusion strategy remarkably enhances the detection ability of IncSAR, resulting in an average accuracy of $99.27\%$, while the last task's accuracy reaches $99.22\%$. This indicates that combining the specialized features produced by the SAR-CNN with the more general features derived by the pre-trained ViT leads to a significant increase in performance. Moreover, when TinyViT \cite{wu2022tinyvit} and SAR-CNN are combined along with the late fusion module the proposed approach derives an average accuracy of $98.63\%$ which is further increased when the late fusion is substituted by the attention module yielding an average accuracy of $99.34\%$, as demonstrated in Table \ref{tab:ablationPretrained}.

\begin{table}[htb]
    \caption{Ablation studies on multiple components of IncSAR on MSTAR dataset.}
    \label{tab:ablationPretrained}
    \centering
    \resizebox{\columnwidth}{!}{%
        \begin{tabular}{lcccccccc}
            Model & SSF  & RPCA & Fusion & RP & LDA & $\bar{A}$ & $A_L$ \\
            \hline
            ViT-B/16 & x & x & x & x & x & $84.63$ & $76.87$ \\
            ViT-B/16 & x & x & x & \checkmark & x & $85.48$ & $77.57$\\
            ViT-B/16 & x & x & x &  x &\checkmark & $98.56$ & $96$\\
            ViT-B/16 & x & x & x & \checkmark & \checkmark & $98.83$ & $96.66$  \\
            ViT-B/16 & \checkmark & x & x & \checkmark & \checkmark & $98.84$ & $97.69$ \\ \hline
            TinyViT \cite{wu2022tinyvit} & x & x & x & \checkmark & \checkmark & $97.53$ & $92.49$ \\
            ViT-Ti \cite{touvron2021training} & \checkmark & x & x & \checkmark & \checkmark & $96.55$ & $91.92$ \\ \hline
            SAR-CNN & x & x & x & \checkmark & \checkmark & $96.45$ & $95.67$  \\
            SAR-CNN & x & \checkmark & x & \checkmark & \checkmark & $98.58$ & $98.14$ \\ \hline
            ViT-B/16 + SAR-CNN & \checkmark & \checkmark & late & \checkmark & \checkmark & $99.27$ & $99.22$ \\ 
            TinyViT + SAR-CNN & \checkmark & \checkmark & late & \checkmark & \checkmark & $98.63$ & $97.77$ \\  
            ViT-Ti + SAR-CNN & \checkmark & \checkmark & attention & \checkmark & \checkmark & $99.34$ & $98.39$ \\  
            \hline
        \end{tabular}%
    } 
\end{table}

\begin{table}[h]
    \caption{Ablation studies on multiple components of IncSAR on SAR-AIRcraft-1.0 dataset and comparisons with state-of-the-art.}
    \label{tab:aircraft}
    \centering
    \resizebox{\columnwidth}{!}{%
        \begin{tabular}{l c c c  c c c c  c  c}  
            \hline
            \multirow{2}{*}{Method} & SSF & RPCA & Fusion & \multicolumn{4}{c}{Accuracy in each task (\%)} & \multirow{2}{*}{PD $\downarrow$} & \multirow{2}{*}{$\bar{A}$ $\uparrow$} \\
            \cline{5-8}
            &  &  &  & $0$ & $1$ & $2$ & $3$  \\
            \hline

            FeCAM\cite{goswami2024fecam} & x & x & x & $77.38$ & $76.90$ & $77.25$ & $77.69$ & $-0.31$ & $77.30$ \\ 
            RanPAC\cite{mcdonnell2024ranpac} & x & x & x & $96.25$ & $95.70$ & $95.83$ & $94.38$ & $1.87$ & $95.54$ \\
            \hline
            ViT-B/16 & x & x & x & $95.75$ & $95.40$ & $95.00$ & $80.92$ & $14.83$ & $91.77$ \\
            ViT-B/16 & \checkmark & x & x & $97.25$ & $97.30$ & $97.08$ & $97.08$ & $0.17$ & $97.18$ \\
            SAR-CNN & x & x & x & $98.88$ & $96.10$ & $98.42$ & $96.00$ & $2.88$ & $97.35$ \\
            SAR-CNN & x & \checkmark & x & $98.88$ & $98.00$ & $97.83$ & $97.00$ & $1.88$ & $97.89$ \\
            ViT-B/16 + SAR-CNN  & \checkmark & x & late & $98.38$ & $97.20$ & $98.17$ & $96.31$ & $2.07$ & $97.52$ \\
            ViT-B/16 + SAR-CNN  & \checkmark & \checkmark & late & $98.38$ & $97.10$ & $97.33$ & $98.31$ & $\mathbf{0.07}$ & $97.78$ \\
            TinyViT \cite{wu2022tinyvit} + SAR-CNN  & x & x & late & $98.88$ & $98.90$ & $98.67$ & $83.31$ & $15.57$ & $94.94$ \\
            TinyViT \cite{wu2022tinyvit} + SAR-CNN & x & \checkmark & late & $98.12$ & $97.70$ & $97.92$ & $97.92$ & $0.20$ & $\mathbf{97.92}$ \\
            ViT-Ti + SAR-CNN & \checkmark & \checkmark & attention & $97.75$ & $97.60$ & $96.83$ & $96.69$ & $1.06$ & $97.21$ \\

            \hline
        \end{tabular}
    }
\end{table}

To further validate the performance of the proposed IncSAR approach, we conducted experiments on the SAR-AIRcraft-1.0 benchmark dataset under the B4Inc1 setup. The selected class order is shown in Table \ref{tab:AIR}. The results shown in Table \ref{tab:aircraft}, highlight the superior performance of IncSAR compared to state-of-the-art methods like FeCAM and RanPAC in both average accuracy and performance drop. The proposed model, which integrates a dual-branch TinyViT \cite{wu2022tinyvit} and SAR-CNN architecture with RPCA and a late fusion strategy, achieves an average accuracy of $97.92\%$, outperforming FeCAM ($77.30\%$) and RanPAC ($95.54\%$) by significant margins. Moreover, IncSAR demonstrates a remarkably low performance drop of $0.20\%$, indicating its strong ability to retain learned knowledge across incremental tasks. In contrast, RanPAC suffers from a PD of $1.87\%$, demonstrating more noticeable degradation in performance as new tasks are introduced. In a variant of the proposed model that includes SSF, RPCA, and the late fusion strategy, the model performs accurately, achieving an average accuracy of $97.78\%$ and an exceptionally low performance drop of $0.07\%$. This minimal PD indicates almost perfect retention of learned knowledge, affirming the effectiveness of both the SSF and RPCA modules in reducing catastrophic forgetting. These components help the model maintain performance stability across all tasks in this challenging incremental learning scenario. When both the SSF and RPCA modules are removed, the model's performance drops sharply, achieving only $94.94\%$ average accuracy, while the performance drop increases drastically to $15.57\%$. This significant degradation highlights the crucial role these components play in both feature extraction and noise reduction in SAR data. Even when only RPCA is removed, the model still maintains strong performance, with an average accuracy of $97.52\%$ and a PD of $2.07\%$. This suggests that the late fusion strategy and SSF continue to contribute to robust performance, though RPCA evidently plays an important role in further reducing PD by denoising the SAR images and improving feature consistency. In another variation of the IncSAR framework, where an attention-based module replaces the late fusion strategy, the model achieves an average accuracy of $97.21\%$ with a performance drop of $1.06\%$. Although this variant slightly underperforms compared to the late fusion approach, the use of the attention mechanism still proves effective in managing task transitions, dynamically weighting important features for improved task-specific learning.

\subsubsection{Comparative analysis of backbone networks}
The detection ability of SAR-CNN within the IncSAR framework is evaluated, comparing its performance against a variety of pre-trained backbone networks. Table \ref{tab:ablation_backbones} demonstrates the comparison of the proposed IncSAR by employing DenseNet-121 \cite{huang2017densely}, ResNet-18 \cite{he2016deep}, ResNet-101 \cite{he2016deep}, VGG-19 \cite{simonyan2014very}, and CLIP-ViT-L/14 \cite{radford2021learning} and the proposed SAR-CNN on the MSTAR dataset under the B2Inc2 setup. The experiments 
maintain the ViT branch remaining consistent, while different networks are tested in the second branch of IncSAR. It is observed that freezing the weights led to better performance compared to fine-tuning them during the base task. SAR-CNN is a lightweight network, that shows remarkable memory efficiency with only $140$k parameters, outperforming the rest of the backbones that require much higher memory budgets. DenseNet-121 requires $7$M parameters, achieving an $\bar{A}$ of $97.92\%$ and a PD of $3.31\%$. When compared to ResNet-101, which yields to $98.37\%$ and $2.47\%$ in  $\bar{A}$ and PD, respectively, SAR-CNN leads to a performance improvement achieving $99.14\%$ in  $\bar{A}$ and   $1.24\%$ in PD outperforming all its competitors. Moreover, CLIP-ViT-L/14 requires $303$M parameters and reaches an  $\bar{A}$ of $98.39\%$ and a PD of $3.1\%$. VGG-19 is lagging behind SAR-CNN, yielding an $\bar{A}$ of $98.30 \%$ and a PD of $ 2.89\%$ and comprising $140$M parameters.

\begin{table}[htb]
    \caption{Comparative analysis of different backbone networks in IncSAR framework.}
    \label{tab:ablation_backbones}
    \centering
    \resizebox{\columnwidth}{!}{%
        \begin{tabular}{l cccc}
             
            Network & Params &  $\bar{A}$ & PD & $A_L$\\
            \hline

            DenseNet-121 \cite{huang2017densely} & $7$M & $97.92$ & $3.31$ & $96.33$ \\
            ResNet-18 \cite{he2016deep} & $11$M & $98.47$ & $2.76$ & $97.24$ \\
            ResNet-101 \cite{he2016deep} & $44$M & $98.37$ & $2.47$ & $97.53$ \\
            VGG-19 \cite{simonyan2014very} & $140$M & $98.30$ & $2.89$ & $97.11$ \\
            CLIP-ViT-L/14 \cite{radford2021learning} & $303$M & $98.39$ & $3.10$ & $96.54$ \\
            \hline
            SAR-CNN & $\mathbf{140}$K & $\mathbf{99.14}$ & $\mathbf{1.24}$ & $\mathbf{98.76}$  \\
            \hline
        \end{tabular}%
    } 
\end{table}

\subsubsection{IncSAR evaluation on limited data scenarios}
 Subsets of the MSTAR dataset are randomly selected to assess the detection ability of the proposed framework under various reduced training data scenarios. Specifically, three different scenarios are tested, employing $80\%$, $50\%$, and $30\%$ of the initial training set. When $50\%$ of the initial training set is employed, IncSAR yields an average accuracy of $98.64\%$, outperforming state-of-the-art MLAKDN and HPECIL methods, which attain $97.96\%$ and  $97.92\%$, respectively. In the challenging scenario of retaining only $30\%$ of samples, IncSAR  demonstrates a performance of $97.48\%$ in terms of average accuracy, which is slightly lower than MLAKDN by $0.48\%$. These results underscore IncSAR's efficiency in detecting SAR images with limited training data, highlighting its capability to generalize well in real-world scenarios. 
 
 Furthermore, we investigate the performance of IncSAR$_{Lite}$ in data-limited scenarios, using the same portion of training data as in previous experiments. When trained with 80\% of the initial training data, IncSAR$_{Lite}$ achieves top performance, with an $\bar{A}$ of $99.7\%$, outperforming MLAKDN, HPecIL, and other IncSAR variants. Additionally, it registers the best  PD  value of $0.45\%$, indicating its robustness in incremental learning, even with reduced training data. Notably, even when the model is trained with 50\% of the initial training set, it achieves a remarkable $\bar{A}$ of $98.99\%$ and maintains a PD of $1.02\%$, outperforming state-of-the-art methods. This demonstrates the effectiveness of the IncSAR$_{Lite}$ variant in scenarios with significantly limited data. In the most challenging case, when only 30\% of the initial training data is used, IncSAR$_{Lite}$ continues to perform impressively, achieving $\bar{A}$ of $97.35\%$, showcasing its adaptability and resilience in extreme data-scarcity conditions. In comparison, the second variant, IncSAR$_{LAtt}$, also delivers strong results. With 80\% of the training data, it achieves an $\bar{A}$ of $98.59\%$, outperforming state-of-the-art methods, though trailing behind the other IncSAR variants. When trained with 50\% of the data, IncSAR$_{LAtt}$ records an $\bar{A}$ of $97.82\%$ with a PD of $3.34\%$, showing good performance but higher forgetting compared to the IncSAR$_{Lite}$ variant. Detailed  results are shown in Table \ref{tab:portion}.

\begin{table}[h]
    \caption{Ablation study of the IncSAR framework under training in different portions of the MSTAR dataset.}
    \label{tab:portion}
    \centering
    \resizebox{\columnwidth}{!}{%
        \begin{tabular}{l cccccccc}  
            \hline
            \multirow{2}{*}{Method} & \multirow{2}{*}{Size (\%)} & \multicolumn{5}{c}{Accuracy in each task (\%)} & \multirow{2}{*}{PD $\downarrow$} & \multirow{2}{*}{$\bar{A}$ $\uparrow$} \\
            \cline{3-7}
            & & $0$ & $1$ & $2$ & $3$ & $4$ \\
            \hline
            \multirow{4}{*}{IncSAR} 
            & $100$ & $100.00$ & $99.89$ & $98.94$ & $99.15$ & $99.22$ & $0.78$ & $99.44$ \\
            & $80$  & $100.00$ & $99.89$ & $98.08$ & $99.09$ & $98.93$ & $1.07$ & $99.19$ \\
            & $50$  & $99.82$ & $99.68$ & $97.94$ & $97.76$ & $98.14$ & $1.68$ & $98.66$ \\
            & $30$  & $100.00$ & $99.47$ & $95.60$ & $95.79$ & $96.54$ & $3.46$ & $97.48$ \\
            \hline

            \multirow{4}{*}{IncSAR$_{Lite}$} 
            & $100$ & $\mathbf{100.00}$ & ${100.00}$ & $99.43$ & ${99.73}$ & $99.38$ & $0.62$ & ${99.70}$ \\
            & $80$  & $100.00$ & $99.89$ & ${99.57}$ & $99.52$ & ${99.55}$ & ${0.45}$ & ${99.70}$ \\
            & $50$  & $99.82$ & $99.89$ & $97.80$ & $98.67$ & $98.80$ & $1.02$ & $98.99$ \\
            & $30$  & $99.64$ & $99.79$ & $96.81$ & $96.91$ & $93.61$ & $6.03$ & $97.35$ \\
            \hline

            \multirow{4}{*}{IncSAR$_{LAtt}$} 
            & $100$ & $100.00$ & $99.89$ & $97.94$ & $97.60$ & $97.90$ & $2.10$ & $98.66$ \\
            & $80$  & $100.00$ & $99.89$ & $97.80$ & $97.28$ & $97.98$ & $2.02$ & $98.59$ \\
            & $50$  & $100.00$ & $99.89$ & $96.45$ & $96.16$ & $96.66$ & $3.34$ & $97.83$ \\
            & $30$  & $100.00$ & $99.36$ & $94.32$ & $93.07$ & $94.10$ & $5.90$ & $96.17$ \\
            \hline

            MLAKDN \cite{yu2023multilevel} & $100$ & $99.64$ & $99.82$ & $98.98$ & $96.87$ & $94.50$ & $5.14$ & $97.96$ \\
            HPecIL \cite{tang2022incremental} & $100$ & $99.45$ & $98.83$ & $98.79$ & $96.70$ & $96.16$ & $3.26$ & $97.92$ \\
            \hline
        \end{tabular}
    }
\end{table}

To further assess the robustness of IncSAR in real-world scenarios with limited training data, we conducted ablation experiments on the SAR-AIRcraft-1.0 benchmark dataset, testing the model with various portions of the training set. The performance of each variant, i.e., IncSAR, IncSAR$_{Lite}$, and IncSAR$_{LAtt}$, was also evaluated at 80\%, 50\%, and 30\% of the initial training data. With the full dataset, IncSAR achieves $\bar{A}$ of $97.78\%$ and a PD of $0.07\%$, showcasing its ability to maintain high accuracy across tasks. When trained with 80\% of the data, IncSAR slightly improves, reaching $\bar{A}$ of $98.23\%$ and a  negative PD of $-0.19\%$, demonstrating stability even with reduced data. As the data availability decreases further, IncSAR’s performance starts to decline. At 50\%, the model reaches a $\bar{A}$ of 95.39\% with an increased PD of 6.11\%, showing some loss in its ability to retain previously learned information. With 30\% of the data, the model achieves $\bar{A}$ of $91.52\%$ with a negative PD of $-0.73\%$, maintaining decent performance but reflecting greater sensitivity to data reduction.

IncSAR$_{Lite}$ achieves noteworthy results even with limited data, demonstrating its flexibility and stability. Using the full 100\% of the data, it achieves $\bar{A}$ of $97.91\%$ and PD of $0.20\%$, comparable to IncSAR’s performance. With 80\% of the data, IncSAR$_{Lite}$ maintains high accuracy, achieving $\bar{A}$ of $97.89\%$ with a PD of $-0.56\%$, again showcasing a small performance gain from the original setup. In lower data regimes, however, IncSAR$_{Lite}$ shows noticeable variance. At 50\% of the data, it achieves $\bar{A}$ of $93.60\%$ and a PD of $-0.04\%$, indicating resilience but with some loss in accuracy. When data availability is reduced to 30\%, the model’s performance declines more significantly, with $\bar{A}$ dropping to $90.09\%$ and PD increasing to $11.58\%$. This highlights that, while IncSAR$_{Lite}$ performs well with moderate data reduction, it becomes more susceptible to performance drops in extreme data-scarce scenarios. IncSAR$_{LAtt}$ also shows good overall performance but is generally outpaced by the other two variants. With the full dataset, IncSAR$_{LAtt}$ achieves $\bar{A}$ of $97.21\%$ and a PD of $1.06\%$, a bit lower than the other two variants. With 80\% of the data, the model’s accuracy decreases to $\bar{A}$ of $94.66\%$ and PD of $0.69\%$, indicating some sensitivity to data reduction. At 50\% data, IncSAR$_{LAtt}$ maintains respectable accuracy at $\bar{A}$ of $94.14\%$ with PD of $0.53\%$. When further reduced to 30\%, it achieves $\bar{A}$ of $88.86\%$ and PD of $0.88\%$, showing effective generalization but a larger drop compared to the other variants. Detailed results are shown in Table \ref{tab:portion_air}.

This evaluation demonstrates that IncSAR and IncSAR$_{Lite}$ perform effectively under reduced data conditions, with IncSAR$_{Lite}$ showing particular resilience at moderate data reductions (80\% and 50\%). IncSAR$_{LAtt}$, while achieving good results, is slightly more impacted by limited data, especially at extreme reductions. These results reinforce the efficacy of the IncSAR framework and its variants in maintaining high accuracy and minimizing catastrophic forgetting across incremental tasks, even in challenging, data-constrained environments.

\begin{table}[t]
    \caption{Ablation study of the IncSAR framework under training in different portions of the SAR-AIRcraft dataset.}
    \label{tab:portion_air}
    \centering
    \resizebox{\columnwidth}{!}{%
        \begin{tabular}{l ccccccc}  
            \hline
            \multirow{2}{*}{Method} & \multirow{2}{*}{Size (\%)} & \multicolumn{4}{c}{Accuracy in each task (\%)} & \multirow{2}{*}{PD $\downarrow$} & \multirow{2}{*}{$\bar{A}$ $\uparrow$} \\
            \cline{3-6}
            & & $0$ & $1$ & $2$ & $3$  \\
            \hline
            \multirow{4}{*}{IncSAR} 
            & $100$ & $98.38$ & $97.10$ & $97.33$ & $98.31$ & $0.07$ & $97.78$ \\
            & $80$  & $98.12$ & $98.50$ & $98.00$ & $98.31$ & $-0.19$ & $98.23$ \\
            & $50$  & $96.88$ & $97.10$ & $96.83$ & $90.77$ & $6.11$ & $95.39$ \\
            & $30$  & $94.12$ & $84.30$ & $92.83$ & $94.85$ & $-0.73$ & $91.52$ \\
            \hline

            \multirow{4}{*}{IncSAR$_{Lite}$} 
            & $100$ & $98.12$ & $97.70$ & $97.92$ & $97.92$ & $0.20$ & $97.91$ \\
            & $80$  & $97.75$ & $98.10$ & $97.42$ & $98.31$ & $-0.56$ & $97.89$ \\
            & $50$  & $96.88$ & $83.70$ & $96.92$ & $96.92$ & $-0.04$ & $93.60$ \\
            & $30$  & $94.50$ & $93.70$ & $89.25$ & $82.92$ & $11.58$ & $90.09$ \\
            \hline

            \multirow{4}{*}{IncSAR$_{LAtt}$} 
            & $100$ & $97.75$ & $97.60$ & $96.83$ & $96.69$ & $1.06$ & $97.21$ \\
            & $80$  & $95.00$ & $95.00$ & $94.33$ & $94.31$ & $0.69$ & $94.66$ \\
            & $50$  & $94.38$ & $94.60$ & $93.75$ & $93.85$ & $0.53$ & $94.14$ \\
            & $30$  & $88.88$ & $89.40$ & $89.17$ & $88.00$ & $0.88$ & $88.86$ \\
            \hline
        \end{tabular}
    }
\end{table}

\subsection{Limitations}

The proposed framework achieves notable improvements over state-of-the-art methods, as demonstrated in the experimental results; however, there are aspects that require further consideration. Although the dual-fusion strategy, which combines a ViT and the custom-designed SAR-CNN is successful in utilizing both global and domain-specific features it is still technically difficult to achieve the optimal trade-off between these branches. Additionally, while improving feature representation the attention-based mechanism in the IncSAR$_{LAtt}$ variant adds architectural complexity and marginally increases computational requirements. Furthermore, even though it is essential for increasing classification accuracy, the added RPCA module for speckle noise reduction results in additional computational overhead, which might make it impractical for applications with limited resources or strict real-time requirements. In order to achieve optimal performance, the framework also depends on hyperparameter tuning such as those pertaining to random projections and LDA decorrelation. This could pose challenges in scenarios involving large distribution shifts that would render the selected hyperparameters suboptimal. Future research could concentrate on resolving these issues by simplifying the framework's architecture to increase scalability and optimize computational efficiency as well as investigating ways to expand its applicability to a wider range of datasets and operational contexts.

\section{Conclusion}
    \label{sec:conclusion}
We proposed an incremental learning framework for SAR target recognition, IncSAR, based on exemplar-free prototype learning. IncSAR integrates a neural network-based RPCA module to reduce SAR speckle noise and employs a random projection layer to improve feature linear separability. Using a late-fusion strategy, IncSAR combines a ViT backbone for generalized features with a specialized custom SAR-CNN for domain-specific details, while an attention-based module enhances feature interactions. IncSAR achieves a strong balance between stability and plasticity, outperforming state-of-the-art methods on MSTAR, SAR-AIRcraft, and OpenSARShip datasets. Extensive evaluations, including data-limited and cross-domain settings, demonstrate IncSAR’s resilience to catastrophic forgetting and robust generalization across SAR domains, supporting its applicability in real-world scenarios.

\bibliographystyle{unsrt}
\bibliography{egbib}

\begin{IEEEbiography}[{\includegraphics[width=1in,height=1.25in,clip,keepaspectratio]{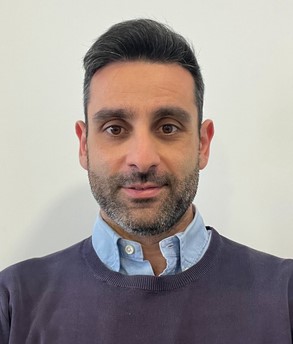}}]{GEORGE KARANTAIDIS} received the Diploma degree in rural and surveying engineering, the Master of Science in computational intelligence and digital media, and the Master of Science in geoinformatics, all from the Aristotle University of Thessaloniki, Greece. He also received the Ph.D. in Signal Processing and Information Analysis at the same institution under the supervision of Prof. Constantine Kotropoulos. Currently, he is a Postdoctoral Research Fellow at the Information Technologies Institute (ITI) of the Centre for Research \& Technology Hellas (CERTH) in Thessaloniki, Greece. He was previously a Ph.D. scholar funded by the Hellenic Foundation for Research and Innovation (HFRI). He is also a member of the Media Analysis, Verification, and Retrieval Group (MeVer) (https://mever.iti.gr). His research interests include deep learning, signal processing, and multimedia forensics.
\end{IEEEbiography}

\begin{IEEEbiography}[{\includegraphics[width=1in,height=1.25in,clip,keepaspectratio]{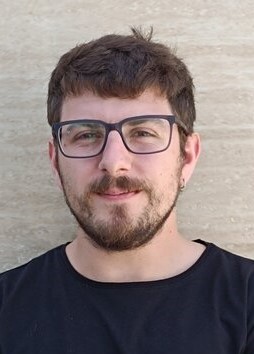}}]{ATHANASIOS PANTSIOS} is currently a Research Assistant in Media Analysis, Verification and Retrieval Group (MeVer) of the Centre for Research \& Technology Hellas (CERTH). He also holds an integrated MSc degree in electrical \& computer engineering from the Aristotle University of Thessaloniki (AUTH). His research interests include Deep Learning, Computer Vision, and Natural Language Processing.
\end{IEEEbiography}

\begin{IEEEbiography}[{\includegraphics[width=1in,height=1.25in,clip,keepaspectratio]{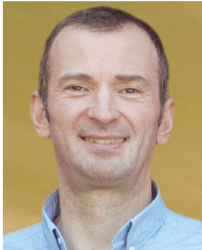}}]{IOANNIS KOMPATSIARIS} (Senior Member, IEEE) is currently the Director of CERTH-ITI and the Head of the Multimedia Knowledge and Social Media Analytics Laboratory. Since January 2014, he has been a Co-Founder of Infalia private company, a high-tech SME focusing on data intensive web services and applications. He is the co-author of 178 articles in refereed journals, 63 book chapters, eight patents, and 560 papers in international conferences. Since 2001, he has been participating in 88 national and European research programs, in 31 of which he has been the project coordinator. He has also been the PI in 15 research collaborations with industry. His research interests include AI/machine learning for multimedia analysis, semantics (multimedia ontologies and reasoning), social media and big data analytics, multimodal and sensors data analysis, human–computer interfaces, e-health, arts and cultural, media/journalism, environmental, and security applications. 

Mr. Kompatsiaris is a member of the National Ethics and Technoethics Committee, the Scientific Advisory Board of the CHIST-ERA Funding Programme and an elected member of the IEEE Image, Video and Multidimensional Signal Processing—Technical Committee (IVMSP—TC). He is a Senior Member of ACM. He has been the Co-Chair of various international conferences and workshops, including the 13th IEEE Image, Video, and Multidimensional Signal Processing (IVMSP 2018) Workshop and has served as a regular reviewer, an associate editor and a guest editor for a number of journals and conferences currently being an Associate Editor of IEEE Transactions on Image Processing.
\end{IEEEbiography}

\begin{IEEEbiography}[{\includegraphics[width=1in,height=1.25in,clip,keepaspectratio]{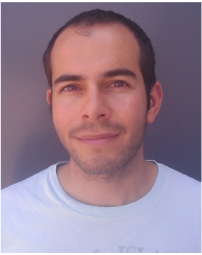}}]{SYMEON PAPADOPOULOS} received the Diploma degree in electrical and computer engineering from the Aristotle University of Thessaloniki, the Professional Doctorate degree in engineering from the Technical University of Eindhoven, the Master of Business Administration degree from the Blekinge Institute of Technology, and the Ph.D. degree in computer science from the Aristotle University of Thessaloniki. He is currently a Principal Researcher with the Information Technologies Institute (ITI), Centre for Research \& Technology Hellas (CERTH), Thessaloniki, Greece. He has co-authored more than 50 articles in refereed journals, 15 book chapters and 150 papers in international conferences, three patents, and has edited two books. His research interests include the intersection of multimedia understanding, social network analysis, information retrieval, big data management, and artificial intelligence. He has participated in and coordinates a number of relevant EC FP7, H2020, and Horizon Europe projects in the areas of media convergence, social media, and artificial intelligence. He is leading the Media Analysis, Verification and Retrieval Group (MeVer) (https://mever.iti.gr) and is a Co-Founder of Infalia Private Company, a spin-out of CERTH-ITI.
\end{IEEEbiography}

\EOD

\end{document}